\documentclass[conference]{IEEEtran}
\usepackage{times}
\usepackage{lipsum}
\usepackage[numbers]{natbib}
\usepackage{multicol}
\usepackage[bookmarks=true]{hyperref}

\usepackage{bbm}
\usepackage{algorithm, algorithmicx, algpseudocode}
\usepackage[caption=false,font=footnotesize]{subfig}
\usepackage{booktabs} 
\usepackage[cmex10]{amsmath}
\usepackage[mathscr]{euscript}
\usepackage{amsfonts}
\usepackage{amssymb}
\usepackage{amsthm}
\usepackage[caption=false,font=footnotesize]{subfig}
\usepackage{fixltx2e}
\usepackage{graphicx}
\usepackage{cleveref}

\usepackage[utf8]{inputenc}
\usepackage[T1]{fontenc}
\usepackage{textcomp}
\usepackage{gensymb}

\usepackage{soul} 
\usepackage{color} 

\newtheorem{problem}{Problem}

\newtheorem{remark}{Remark}
\newtheorem{definition}{Definition}
\newtheorem{example}{Example}
\newtheorem{assumption}{Assumption}

\newcommand{\Ccal}{\mathcal{C}}
\newcommand{\Dcal}{\mathcal{D}}
\newcommand{\Lcal}{\mathcal{L}}
\newcommand{\Mcal}{\mathcal{M}}
\newcommand{\Ncal}{\mathcal{N}}
\newcommand{\Ocal}{\mathcal{O}}

\newcommand{\Xcal}{\mathcal{X}}

\newcommand{\Zcal}{\mathcal{Z}}

\newcommand{\m}{\mathop{\mathrm{m}}}

\newcommand{\bigO}[1]{\Ocal(#1)} 
\newcommand{\EV}[1]{\mathbb{E}[#1]} 
\newcommand{\Var}[1]{\mathbb{V}[#1]} 

\begin{document}

\title{Gaussian Processes Semantic Map Representation}

\author{\authorblockN{Maani Ghaffari Jadidi, Lu Gan, Steven A. Parkison, Jie Li, and Ryan M. Eustice}
\authorblockA{Perceptual Robotics Laboratory, Department of Naval Architecture and Marine Engineering\\
University of Michigan, Ann Arbor, MI 48109 USA\\
{\tt \{{maanigj, ganlu, sparki, ljlijie, eustice\}@umich.edu}}}}

\maketitle

\begin{abstract}
In this paper, we develop a high-dimensional map building technique that incorporates raw pixelated semantic measurements into the map representation. The proposed technique uses Gaussian Processes (GPs) multi-class classification for map inference and is the natural extension of GP occupancy maps from binary to multi-class form. The technique exploits the continuous property of GPs and, as a result, the map can be inferred with any resolution. In addition, the proposed GP Semantic Map (GPSM) learns the structural and semantic correlation from measurements rather than resorting to assumptions, and can flexibly learn the spatial correlation as well as any additional non-spatial correlation between map points. We extend the OctoMap to Semantic OctoMap representation and compare with the GPSM mapping performance using NYU Depth V2 dataset. Evaluations of the proposed technique on multiple partially labeled RGBD scans and labels from noisy image segmentation show that the GP semantic map can handle sparse measurements, missing labels in the point cloud, as well as noise corrupted labels.

\end{abstract}

\IEEEpeerreviewmaketitle

\section{Introduction}

Semantic knowledge in robotic perception systems can use representations such as hierarchical maps~\citep{kuipers1991robot}, objects as higher level landmarks~\citep{salas2013slam++}, or voxelized reconstruction~\citep{kochanov2016scene}. Dense robotic maps such as occupancy grids and the \mbox{OctoMap}~\citep{moravec1985high,elfes1987sonar,thrun2003learning,hornung2013octomap,merali2014optimizing} traditionally contain geometric knowledge of the environment. Grid/voxel-based maps have been successful in many applications such as localization, robotic exploration, and navigation tasks~\citep{yamauchi1997frontier,stachniss2005information,charrow2015information,wolcott2017robust}. However, these techniques ignore available correlations in data by simplifying the mapping problem into a set of marginalized random variables. Furthermore, the map resolution is often fixed or limited and once the map is inferred the resolution cannot be increased. In the context of high-dimensional occupancy mapping, at the cost of higher computational time, \emph{Gaussian Processes} (GPs) have improved the map building performance by taking into account the correlation between map points and treating the map inference as a \emph{binary classification} problem~\citep{t2012gaussian,kim2013occupancy,maani2014com,maani2017auro}.

Semantic segmentation of the scene has long been an active topic in computer vision. The best performing algorithms used to rely on classifiers trained on a set of hand-crafted features~\cite{brostow2008segmentation,shotton2008semantic}. However, the computational efficiency limits their application in real-time mobile robotics scenarios.
More recent literature use the advances in the \emph{deep convolutional neural network}. New architectures designed for semantic segmentation, including~\cite{long2015fully,badrinarayanan2015segnet,chen2016deeplab,noh2015learning}, have achieved superior performance in both indoor and outdoor benchmarks. Furthermore, the processing time for their pixel-wise estimation is also promising.

\begin{figure}[t]
\vspace{.25cm}
  \centering 
  \subfloat{\includegraphics[width=.5\columnwidth,trim={0cm 0cm 0cm 0cm},clip]{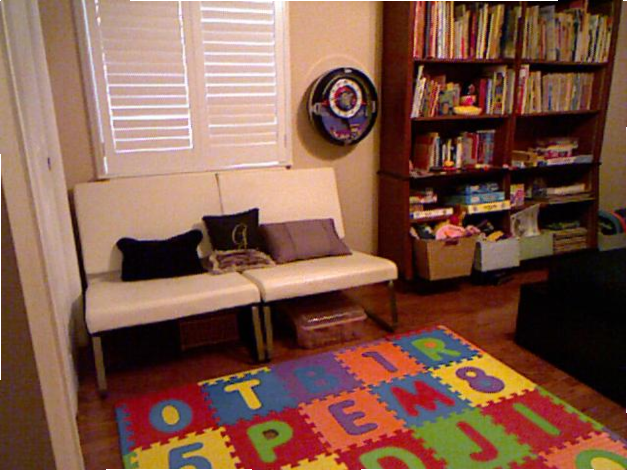}
  \label{fig:gt}}
  \subfloat{\includegraphics[width=.5\columnwidth,trim={0cm 0cm 0cm 0cm},clip]{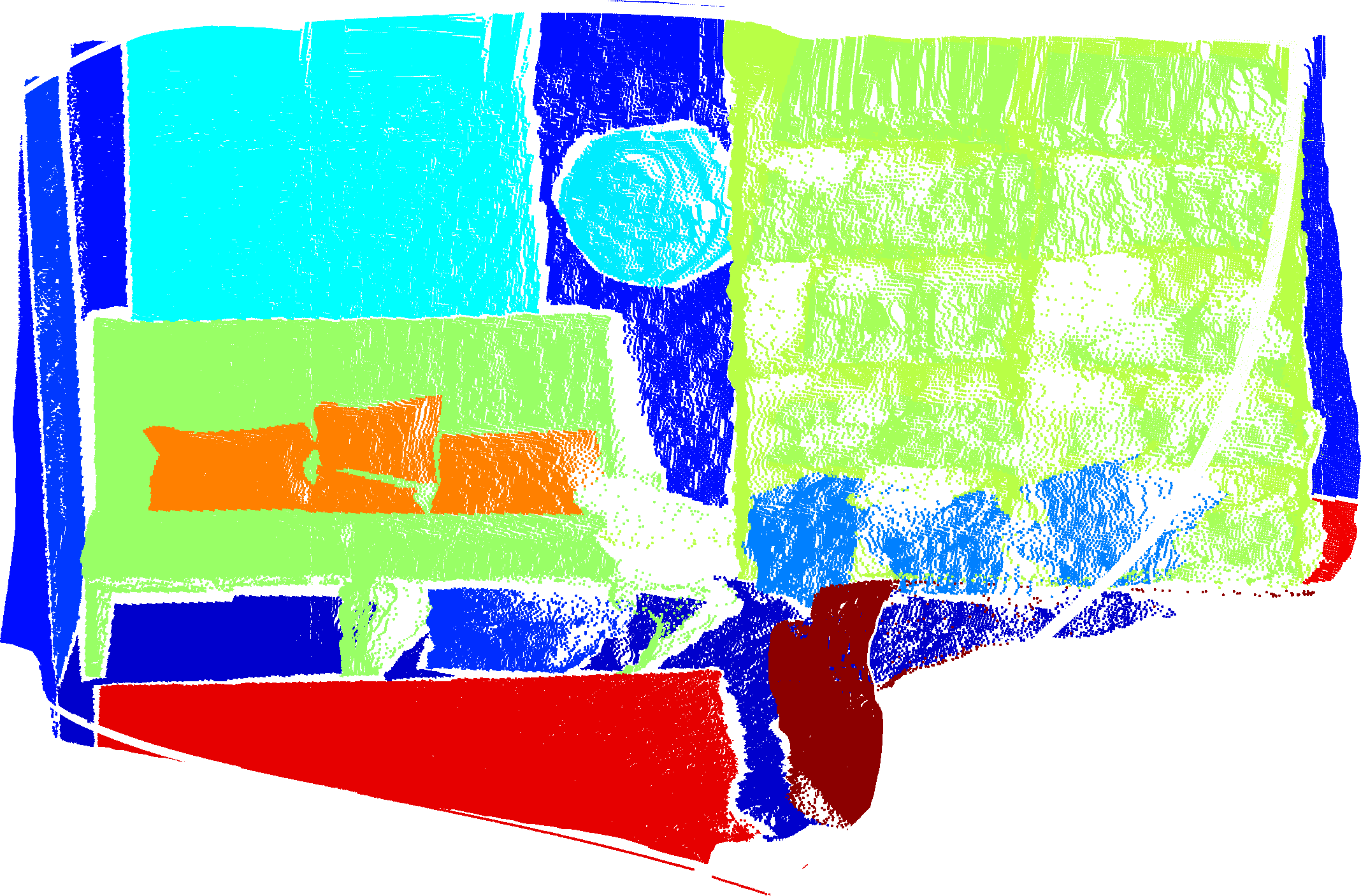}
  \label{fig:gtd}} \\
  \subfloat{\includegraphics[width=.5\columnwidth,trim={0cm 0cm 0cm 0cm},clip]{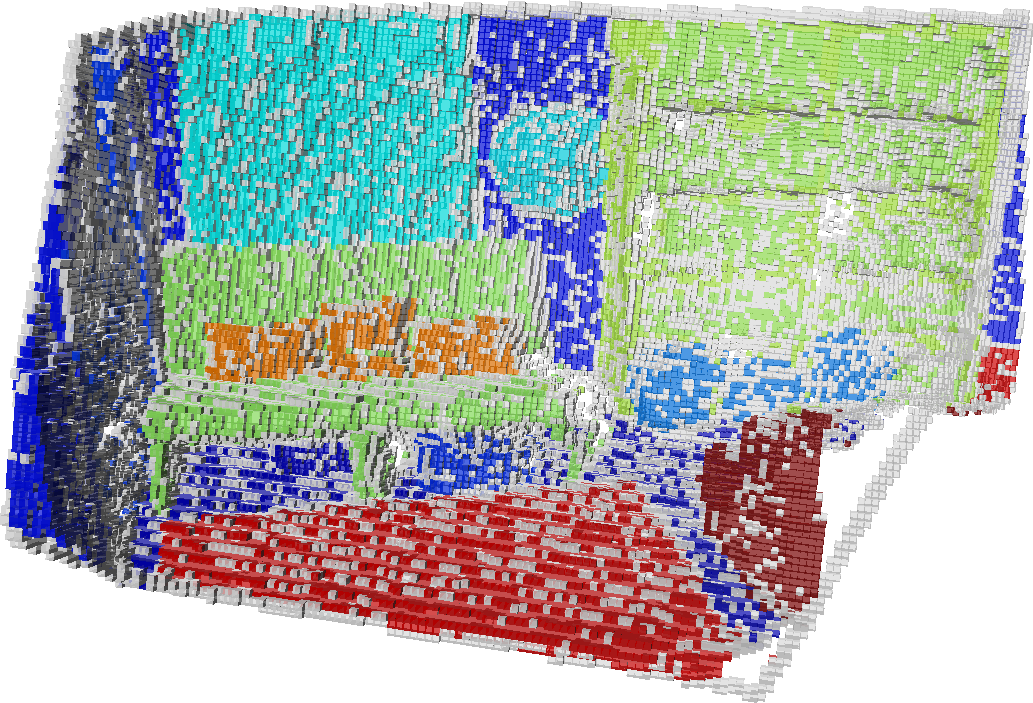}
  \label{fig:som}}
  \subfloat{\includegraphics[width=.5\columnwidth,trim={0cm 0cm 0cm 0cm},clip]{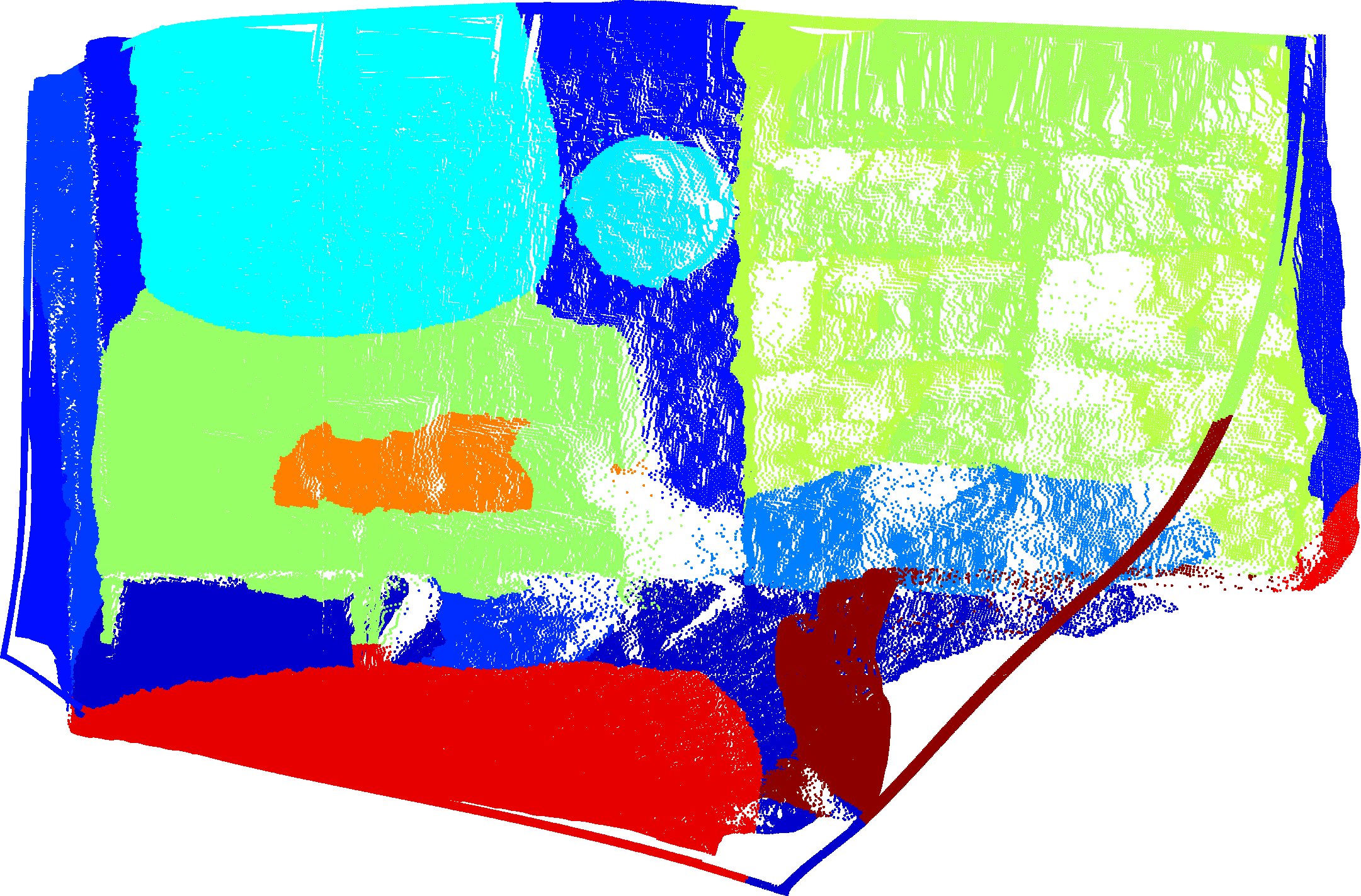}
  \label{fig:gpsm}} 
  \caption{The figure shows examples of the GPSM (bottom right) and SOM (bottom left) using NYU Depth V2 dataset. The point cloud with ground truth semantic labels is shown in the top right. The maps are built by uniformly down-sampling the original point cloud to one-third. The GP map can infer the missing labels and deal with sparse measurements. The query points are the same as original points in the top right.}
  \label{fig:firstfig}
\end{figure}

In this paper, we formulate the semantic mapping as a \emph{multi-class classification} problem and use raw pixelated semantic measurements (class labels) to generalize the traditional occupied and unoccupied class assignment. The proposed Gaussian Processes Semantic Map (GPSM) is inherently continuous, and prediction can be made at any desired location. We use \emph{kernel methods} in the form of GPs to systematically accept inputs with any dimensionality as well as heterogeneous bases such as spatial coordinates and colors. Figure~\ref{fig:firstfig} shows examples of Semantic OctoMap (SOM) and GPSM built using a single frame of NYU Depth V2 dataset~\citep{Silberman:ECCV12}. In particular, this work has the following contributions:
\begin{enumerate}
\item The proposed technique infers the structural and semantic correlation from measurements rather than resorting to assumptions. Through learning the correlation in data, GPSM can infer missing labels and deal with sparse measurements.
\item GPSM is continuous, and queries can be made at any desired locations; therefore, the map can be inferred with any resolution.
\item GPSM is the extension of GPs occupancy maps, i.e.\@ binary maps, to the multi-class semantic representation which provides rich maps for robotic planning tasks.
\item GPSM is agnostic to the input dimensions and can handle an arbitrary number of non-spatial dimensions~\footnote{Without loss of generality, in this work, we only use spatial inputs to compare the results with semantic OctoMap.}.
\end{enumerate}

\subsection*{Outline}
In the following section, a review of the related work is given. The problem statement and preliminaries are discussed in Section~\ref{sec:prob_state}. The detailed formulation of the GP semantic mapping is presented in Section~\ref{sec:gpsm}. The semantic OctoMap as an alternative map building technique is described in Section~\ref{sec:octomap}. Section~\ref{sec:complexity} includes the time complexity analysis of both GP semantic map and semantic OctoMap. The Comparison of mapping results using a publicly available dataset is presented in Section~\ref{sec:results}; and finally, Section~\ref{sec:conclusion} concludes the paper and discusses possible extensions of this work.

\subsection*{Notation}
\label{subsec:notation}
In the present article probabilities and probability densities are not distinguished in general. Matrices are capitalized in bold, such as in $\boldsymbol X$, and vectors are in lower case bold type, such as in $\boldsymbol x$. Vectors are column-wise and $1\colon n$ means integers from $1$ to $n$. The Euclidean norm is shown by $\lVert \cdot \rVert$. $\lvert \boldsymbol X \rvert$ denotes the determinant of matrix $\boldsymbol X$. For the sake of compactness, random variables, such as $X$, and their realizations, $x$, are sometimes denoted interchangeably where it is evident from context. $x^{[i]}$ denotes a reference to the $i$-th element of the variable. An alphabet such as $\mathcal{X}$ denotes a set, and the cardinality of the set is denoted by $\lvert \mathcal{X} \rvert$. A subscript asterisk, such as in $\boldsymbol x_*$, indicates a reference to a test set quantity. The $n$-by-$n$ identity matrix is denoted by $\boldsymbol I_{n}$. We use $\mathrm{vec}(\boldsymbol x, \boldsymbol y)$ to construct a vector by stacking $\boldsymbol x$ and $\boldsymbol y$. The function notation is overloaded based on the output type and denoted by $k(\cdot)$, $\boldsymbol k(\cdot)$, and $\boldsymbol K(\cdot)$ where the outputs are scalar, vector, and matrix, respectively. Finally, $\EV\cdot$ and $\Var\cdot$ denote the expected value and variance of a random variable, respectively.

\section{Related Work}
\label{sec:relatedwork}

Early work in the context of robotic semantic mapping focused on room and building structure semantics, similar to topological mapping. \citet{kuipers1991robot} create a network of distinctive places to map and explore large scale environments. \citet{mozos2007supervised} assign labels to a 2D map of the indoor space corresponding to different parts of an indoor environment, such as room, corridor, and doorway through applying a classifier to the information collected by a range sensor. Then a \emph{Hidden Markov Model} is used to encode transition probabilities to different labels, and thus using topological information. \citet{pronobis2010multi} extend that framework to include visual features, namely \emph{Scale-Invariant Feature Transform} (SIFT)~\cite{lowe2004distinctive} and \emph{Composite Receptive Field Histograms} (CRFH)~\cite{linde2004object}.
Another interesting form of semantics for \emph{Simultaneous Localization And Mapping} (SLAM) is object level classification. While room level semantics and topology are useful for navigation and exploration, object level semantics enable finer grained tasks such as \emph{robotic manipulation}. \citet{castle2007towards} use a database of known planar objects to match SIFT features of a monocular video stream in an \emph{extended Kalman filter} SLAM framework. \citet{civera2011towards} extend this approach to arbitrary three-dimensional (3D) geometries. \citet{bao2011semantic} use an object detector in the structure from motion setup to jointly estimate camera parameters, 3D points, and object instances and poses. These approaches are developed to improve scene estimation by providing more geometric constraints. Conversely, \citet{pillai_rss15} use monocular SLAM to aggregate multiple views of a single object to provide more evidence to the object detector.

Dense 3D priors of objects have also been used for mapping and scene understanding. \citet{kim2012acquiring} learn models that are a collection of primitive 3D shapes; and show once a model based on primitives is learned, the object can quickly be recognized in an environment. \citet{salas2013slam++} align 3D mesh model priors of objects to the RGBD frame. The technique treats objects as landmarks and each alignment as a \emph{factor} in the \emph{graphical} SLAM framework. \citet{choudhary2014slam} also use objects as landmarks, but instead of having a dense 3D prior for every object, the objects are discovered via segmentation, and then their models are produced during the mapping process. 

Beyond just object models, there has been work to produce 3D maps with dense semantic labels. \citet{kundu2014joint} jointly reconstruct the 3D scene and perform semantic segmentation. The technique uses a \emph{Conditional Random Field} (CRF) to infer the semantic category and occupancy for each voxel jointly. \citet{sengupta2015semantic} also look at semantic segmentation and reconstruction of the 3D scene; the technique uses stereo images, estimated camera pose, and a CRF defined over voxels and \emph{supervoxels} of the \emph{octree} to infer the semantic octree representation of the 3D scene. \citet{vineet2015incremental} propose an incremental dense stereo reconstruction and semantic segmentation technique. To address the challenge of dealing with moving objects, \citet{kochanov2016scene} present a method to incorporate temporal updates into the map using scene flow measurements.

In this paper, we propose the GP semantic map to infer the structural and semantic representation of the scene concurrently. Our approach fundamentally differs from the above-mentioned techniques as it does not discretize the map and is continuous. While in the present work, we focus on the problem formulation and 3D reconstruction of a single RGBD scan, the proposed framework can be used for the 3D scene reconstruction using map fusion algorithms~\citep{maani2017auro}. Furthermore, the techniques mentioned earlier ignore pose estimation uncertainty in the map building process; however, our framework is systematically capable of accepting uncertain inputs~\citep{maaniwgpom}.

\section{Problem Statement and Preliminaries}
\label{sec:prob_state}

The objective is formulate and solve the mapping process in a fully probabilistic framework. Since the environment representation we consider here is dense and measurements can be sparse due to the limited sensor field of view and range, a simple inference on individual voxels can lead to a poor mapping performance. Therefore, we devise a joint inference scheme based on GPs by assuming map points are normally distributed. In particular, the following assumption are made.

\begin{assumption}[Static map representation]
 The environment is static.
\end{assumption}
\begin{assumption}[Gaussian map points]
\label{assump:gppoint}
 Any sampled point from the semantic map representation of the environment is a random variable whose distribution is Gaussian.
\end{assumption}

From Assumption~\ref{assump:gppoint}, and placing a joint distribution over map points, the mapping process by definition can be modeled as a GP.

\begin{definition}[Gaussian process~\citep{rasmussen2006gaussian}]
A Gaussian process is a collection of random variables, any finite number of which have a joint Gaussian distribution.
\end{definition}
\begin{problem}[Gaussian processes semantic map]
\label{prob:gpsm}
Given a point cloud measurement that is (possibly partially) assigned with noisy semantic class labels, infer a semantic map representation of the point cloud as a Gaussian process.
\end{problem}

\subsection{Gaussian Processes Regression}
\label{subsec:gpreg}
GPs are nonparametric Bayesian regression techniques that employ statistical inference to learn dependencies between points in a data set~\citep{rasmussen2006gaussian}. The joint distribution of the observed \emph{target} values, $\boldsymbol y$, and the function values (the latent variable), $\boldsymbol f_*$, at the query points can be written as
\begin{equation}
\label{eq:gp_joint}
 \begin{bmatrix}
	\boldsymbol y \\
	\boldsymbol f_*
 \end{bmatrix} \sim \mathcal{N}(\boldsymbol 0,
 \begin{bmatrix}
	\boldsymbol K(\boldsymbol X,\boldsymbol X)+\sigma_n^2 \boldsymbol I_{n} & \boldsymbol K(\boldsymbol X,\boldsymbol X_*) \\
	\boldsymbol K(\boldsymbol X_*,\boldsymbol X)			& \boldsymbol K(\boldsymbol X_*,\boldsymbol X_*) 
 \end{bmatrix})
\end{equation}
where $\boldsymbol X$ is the $d\times n$ design matrix of aggregated input vectors $\boldsymbol x$, $\boldsymbol X_*$ is a $d\times n_*$ query points matrix, $\boldsymbol K(\cdot,\cdot)$ is the GP covariance matrix, and $\sigma_n^2$ is the variance of the observation noise which is assumed to have an independent and identically distributed (i.i.d.) Gaussian distribution. The predictive conditional distribution for a single query point $f_*|\boldsymbol X,\boldsymbol y,\boldsymbol x_* \sim \mathcal{N}(\EV{f_*},\Var{f_*})$ can be derived as 
\begin{equation}
 \label{eq:gp_mean}
 \EV{f_*} = \boldsymbol k(\boldsymbol X,\boldsymbol x_*)^{T}[\boldsymbol K(\boldsymbol X,\boldsymbol X)+\sigma_n^2 \boldsymbol I_{n}]^{-1}\boldsymbol y
\end{equation}
\begin{align}
\label{eq:gp_cov}
 \nonumber \Var{f_*} = &k(\boldsymbol x_*,\boldsymbol x_*) \\
 &- \boldsymbol k(\boldsymbol X,\boldsymbol x_*)^{T}[\boldsymbol K(\boldsymbol X,\boldsymbol X)+\sigma_n^2 \boldsymbol I_{n}]^{-1}\boldsymbol k(\boldsymbol X,\boldsymbol x_*)
\end{align}

\subsection{Gaussian Processes Classification}
\label{subsec:gpc}
\emph{Supervised classification} is the problem of learning input-output mappings from a training dataset for discrete outputs (class labels). In binary GP Classification (GPC), we define class labels as $y \in \{\pm1\}$. In GPC, the inference is performed in two steps. First computing the predictive distribution of the latent variable corresponding to a query case, $f_*|\boldsymbol X,\boldsymbol y,\boldsymbol x_* \sim \mathcal{N}(\mathbb{E}[f_*],\mathbb{V}[f_*])$, and then a probabilistic prediction, $p(y_*=+1|\boldsymbol X,\boldsymbol y,\boldsymbol x_*)$, using a \emph{sigmoid function}, $\sigma(\cdot)$, that assigns class labels with a probability that increases monotonically with the latent. The non-Gaussian likelihood and the choice of the sigmoid function can make the inference analytically intractable. Hence, approximate inference techniques are required.

\subsection{Model Selection}
\label{subsec:gpmodel}
In GPs, one can learn free parameters of mean, covariance, and likelihood functions (\emph{hyperparameters}) through optimizing a cost function. Practically, The hyperparameters, $\boldsymbol\theta$, can be computed by minimization of the negative log of the marginal likelihood (NLML) function.
\begin{align}
\label{eq:nlml}
	\nonumber \log p(\boldsymbol y|\boldsymbol X,\boldsymbol\theta) &= -\frac{1}{2}\boldsymbol y^{T}(\boldsymbol K(\boldsymbol X,\boldsymbol X)+\sigma_n^2 \boldsymbol I_{n})^{-1}\boldsymbol y \\
	&-\frac{1}{2}\log \arrowvert K(\boldsymbol X,\boldsymbol X)+\sigma_n^2 \boldsymbol I_{n} \arrowvert-\frac{n}{2}\log 2\pi\
\end{align}
In~\eqref{eq:nlml}, the first term $-\frac{1}{2}\boldsymbol y^{T}(\boldsymbol K(\boldsymbol X,\boldsymbol X)+\sigma_n^2 \boldsymbol I_{n})^{-1}\boldsymbol y$ corresponds to data-fit, the second term $\frac{1}{2}\log\ \arrowvert K(\boldsymbol X,\boldsymbol X)+\sigma_n^2 \boldsymbol I_{n} \arrowvert$ penalizes the model complexity, and the last term is a constant.
\begin{remark}
 Note that the selection of NLML function for the optimization problem is entirely optional. However due to the marginalization property of the multivariate normal distribution, the latent variable $\boldsymbol f_*$ can be conveniently marginalized.
\end{remark}
\begin{remark}
Generally speaking, building maps using GPs can handle sparse sensor observations and consequently sparse training data. However, in practice, the kernel function describes the correlation between training points. A smooth kernel such as the squared exponential can cover a larger area with fewer training points, and a rough kernel such as Mat\'ern ($\nu=1/2$) can only cover the vicinity of sparse training points, see~\citet[Figure 4.1]{rasmussen2006gaussian}. Therefore, model selection is crucial to fully exploit capabilities of GPs. Furthermore, hyperparameters have a significant effect on the shape of the map.
\end{remark}

The GPC model implemented in this work uses a constant mean function, Mat\'ern ($\nu=5/2$) covariance function~\citep{stein1999interpolation} with automatic relevance determination~\citep{neal1996bayesian}, the error function likelihood (\emph{probit regression}), and \emph{Laplace} technique for approximate inference and is done using the open source GP library in~\cite{rasmussen2006gaussian}. It is argued in~\citet{stein1999interpolation} that Mat\'ern covariance functions are more suitable for modeling physical processes.

\subsection{Large-scale Inference}
\label{subsubsec:gpinf}
The main bottleneck in GP regression is computation of the term $[\boldsymbol K(\boldsymbol X,\boldsymbol X)+\sigma_n^2 \boldsymbol I_{n}]^{-1}$ where the covariance matrix of training data has to be inverted. This limitation reveals itself more when one tries to use a large number of training data. In general, the number of measurements in a point cloud exceeds a few hundred and can be up to several hundred thousands. In such cases, approximate inference techniques such as Laplace, \emph{expectation propagation}~\citep{minka2001family}, or \emph{variational Bayes}~\citep{jordan1999introduction} can become time-consuming. The \emph{Fully Independent Training Conditional} (FITC)~\citep{snelson2006sparse,naish2007generalized} method is based on a low-rank plus diagonal approximation to the exact covariance matrix and is computationally more attractive while it preserves the desirable properties of the full GP~\citep{snelson2006sparse}. In particular, FITC uses a set of \emph{inducing points} to shift the computational cost on the \emph{cross-covariances} computation between training, test, and inducing points.

\section{Gaussian Processes Semantic Map}
\label{sec:gpsm}
In this section, we formulate the GP semantic map to solve Problem~\ref{prob:gpsm}. While the formulation we present here is general, for the classification purpose, we use a binary GPC as the base classifier for each class and \emph{one-vs.-rest} approach to building a multi-class classifier. However, we acknowledge that the \emph{multi-class Laplace} approximation in~\citet[Section 3.5]{rasmussen2006gaussian} is an interesting approach to building a true probabilistic multi-class classifier.

Let $\Mcal$ be the set of possible semantic maps. We consider the map of the environment as an $n_m$-tuple random variable \mbox{$(M^{[1]},\dots,M^{[n_m]})$} whose elements are described by a normal distribution \mbox{$m^{[i]} \sim \mathcal{N}(\mu^{[i]},v^{[i]})$, $i \in \{1\colon n_m\}$}. Let $\Xcal \subset \mathbb{R}^3$ be the set of spatial coordinates to build a map on, and $\Ccal=\{c^{[j]}\}_{j=1}^{n_c}$ be the set of semantic class labels. Let \mbox{$\Zcal \subset \Xcal \times \Ccal$} be the set of possible measurements. The observation consists of an $n_z$-tuple random variable $(Z^{[1]},\dots,Z^{[n_z]})$ whose elements can take values \mbox{$\boldsymbol z^{[k]} \in \Zcal$, $k \in \{1\colon n_z\}$} where $\boldsymbol z^{[k]} = (\boldsymbol x^{[k]},y^{[k]})$, $\boldsymbol x^{[k]} \in \Xcal$, and $y^{[k]} \in \Ccal$.

We define a training set \mbox{$\Dcal = \{(\boldsymbol x^{[i]},y^{[i]})\}_{i=1}^{n_t}$} and the target vector \mbox{$\boldsymbol y=\mathrm{vec}(y^{[1]},\dots,y^{[n_t]})$} where $\Dcal \subseteq \Zcal$, and $n_t \leq n_z$ is the number of training points. Given observations $Z= \boldsymbol z$, we wish to estimate \mbox{$p(M=m\mid Z= \boldsymbol z)$}. The map can be inferred as a Gaussian process by defining the process as the function $y:\Zcal\rightarrow\Mcal$, therefore
\begin{equation}
 \label{eq:mapGP}
 y(\boldsymbol x) \sim \mathcal{GP}(f_m(\boldsymbol x), k(\boldsymbol x,\boldsymbol x'))
\end{equation}
It is often the case that we set the mean function $f_m(\boldsymbol x)$ as zero, unless it is mentioned explicitly that $f_m(\boldsymbol x)\neq0$. For a given query point in the map, $\boldsymbol x_*$, GP predicts a mean, $\mu$, and an associated variance, $v$. Thus, for any map point, we have  $m^{[i]} = y(\boldsymbol x^{[i]}_*) \sim \mathcal{N}(\mu^{[i]},v^{[i]})$.

Once the mean and variance of the latent $f_*$ are available, we predict the averaged predictive probability of the class $c^{[j]}$ as follow~\citep[Chapter 3]{rasmussen2006gaussian}.
\begin{equation}
 \label{eq:gpclasspredict}
 p(c^{[j]}=+1|\Dcal,\boldsymbol x_*) = \int \sigma(u)\Ncal(u\mid \EV{f_*}, \Var{f_*}) du
\end{equation}

Note that once the $n_c$ binary GPC are trained, and the prediction at query points (map points) are performed, we normalize the class probabilities to get $p(M=c^{[j]}|\boldsymbol z)$ for the $j$-th semantic class. One straightforward way to assign hard labels to map points is to find the class with the maximum probability.

\begin{remark}
 The actual representation of the map depends on the distribution of query points. It is often the case to use uniformly distributed points. In general, query points can have any desired distributions. However, in this work, we use the original dense point cloud as the query points to facilitate comparison with the ground truth.
\end{remark}

\begin{figure}[t]
  \centering 
  \subfloat{\includegraphics[width=.5\columnwidth,trim={0cm 0cm 0cm 0cm},clip]{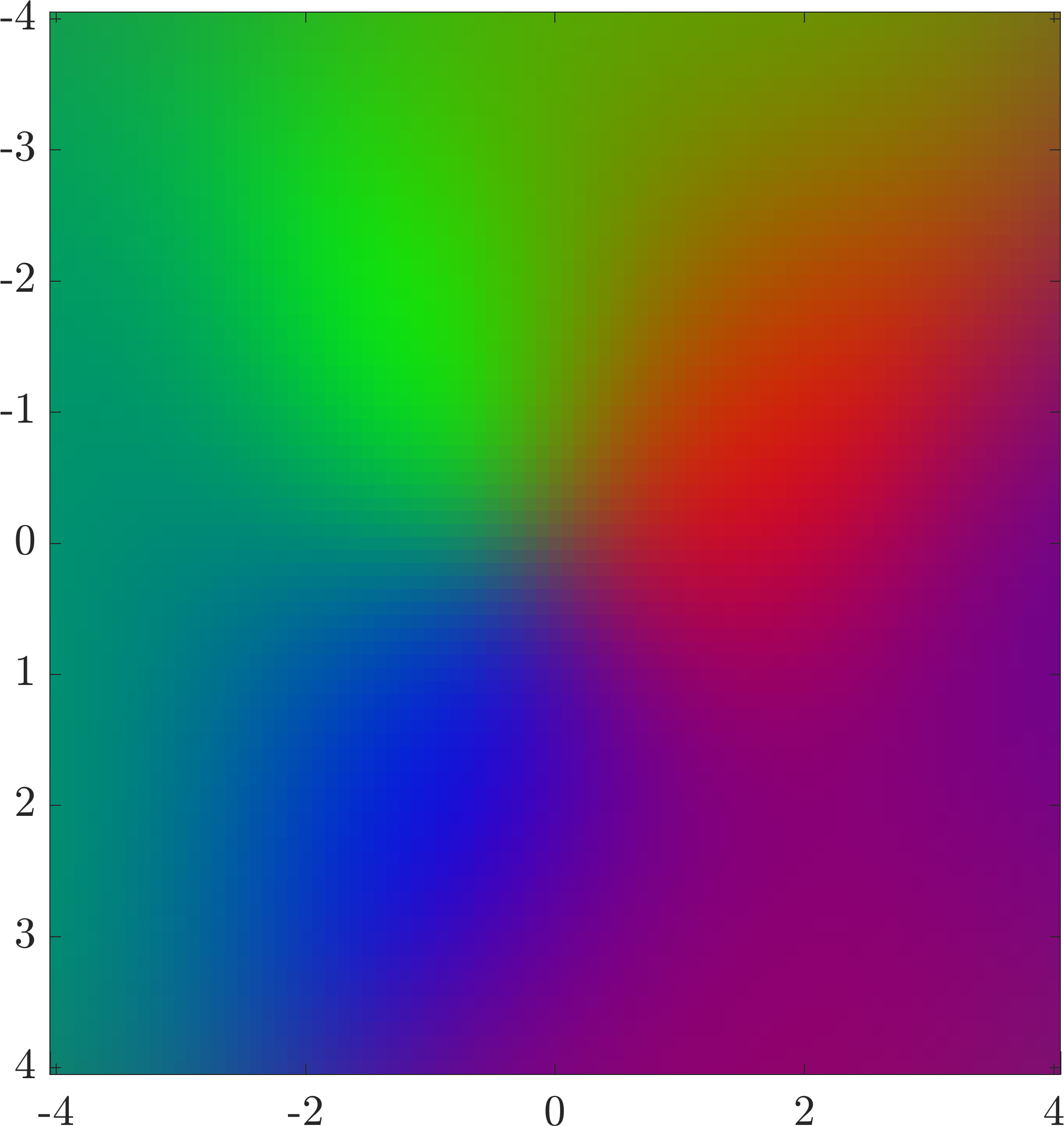}
  \label{fig:toygpc}}
  \subfloat{\includegraphics[width=.5\columnwidth,trim={0cm 0cm 0cm 0cm},clip]{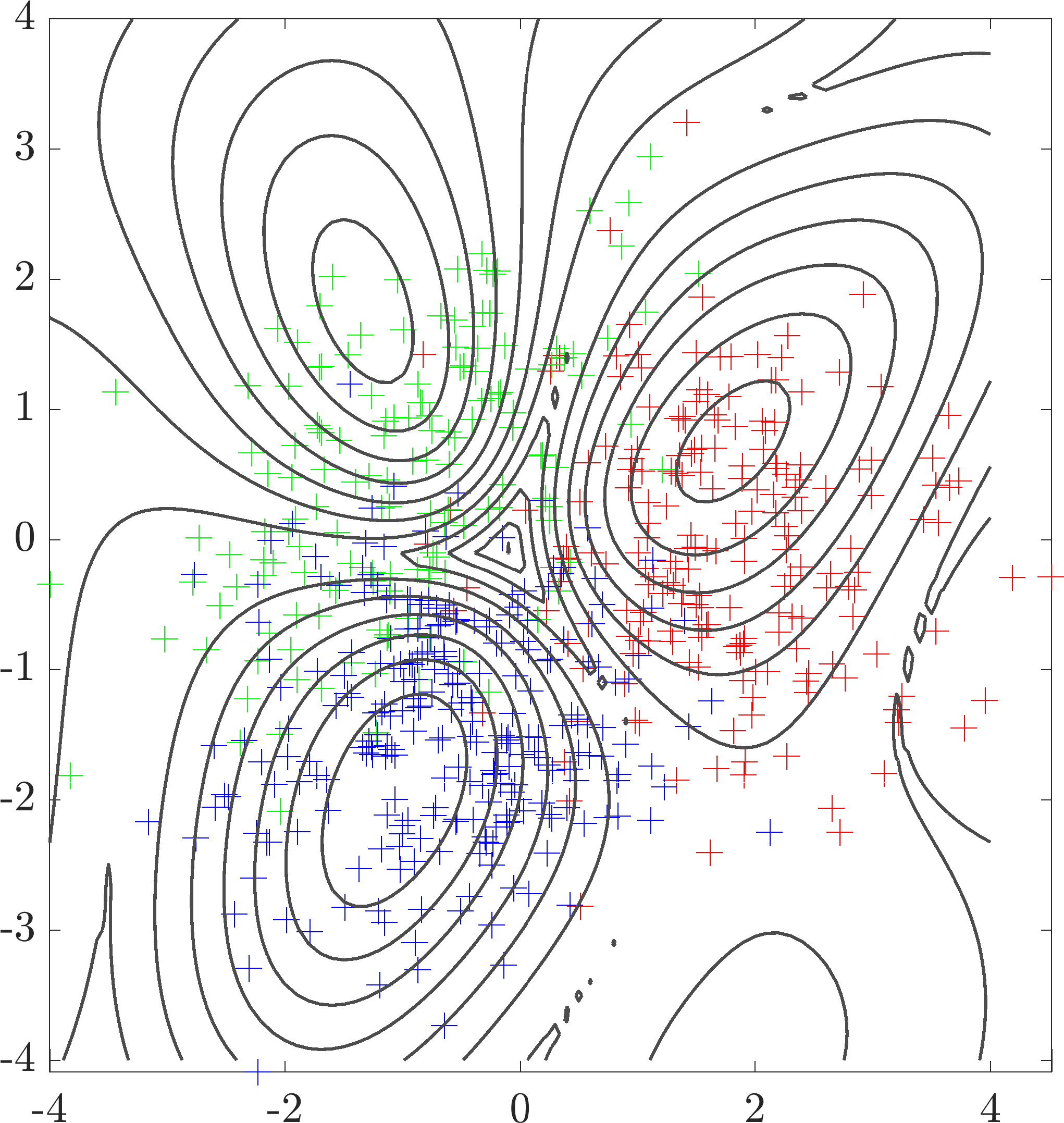}
  \label{fig:toygpccon}} 
  \caption{A two-dimensional toy example of GP multi-class classification using the synthetic dataset shown in the right plot. The prediction is continuous, and the class probabilities can be queries at any point in the plane as shown in the left plot. Furthermore, the contour curves can be seen as decision boundaries. The plots show the log probabilities.}
  \label{fig:toymap}
\end{figure}

\begin{example}[Two-dimensional Toy Example]
Figure~\ref{fig:toymap} illustrates a two-dimensional toy example of GP multi-class classification that can be extended to the three-dimensional semantic mapping. The continuity and smoothness of the probabilistic inference are evident from the left plot, while the corresponding contour curves and the synthetic training points are shown in the right plot. The plots show the log probabilities.
\end{example}

\section{Semantic OctoMap}
\label{sec:octomap}

The OctoMap~\citep{hornung2013octomap} is a popular robotic occupancy mapping tool which builds a discretized model of the 3D world using the octree data structure. The octree enables the multi-resolution map representation which has the advantage of being memory-efficient. In this section, we develop a simple extension of this technique to semantic OctoMap which will serve as the alternative approach in the presented evaluations in Section~\ref{sec:results}. In the basic OctoMap implementation, voxels only contain occupancy probability represented as log-odds. In the developed semantic OctoMap, in addition, voxels include semantic labeling and color information for visualization. The incorporation of the semantic knowledge into the OctoMap is also discussed in~\citet{sengupta2015semantic}.

Following the problem formulation in the previous section, given observations $Z= \boldsymbol z$, we wish to estimate \mbox{$p(M=m\mid Z= \boldsymbol z)$}. The fundamental assumptions of this approach are as follows. First, the map posterior is approximated as the product of its marginals \cite{thrun2005probabilistic}:

\begin{equation}
 \label{eq:mapy}
p(M=m\mid Z= \boldsymbol z) = \prod_{i=1}^{n_m} p(M = m^{[i]}\mid Z =  \boldsymbol z)
\end{equation}
which indicates the distribution of each voxel is independent of the others. Second, we assume that for each voxel, the occupancy and semantic probabilities are independent. Based on these tow assumptions, we can process occupancy and semantic beliefs separately. We first use OctoMap library to compute the occupancy probability of each voxel. The principle is that voxels correspond to endpoints are updated as hits, while voxels along the ray between the sensor origin and the endpoint are updated as misses. 

We follow the idea in~\citet{kochanov2016scene} to define the semantic likelihood. The idea is to simply average the discrete semantic class labels that fall within a voxel. Let $\Lcal^{[i]}$ be a multiset that contains semantic label observation, $l^{[i,j]} \in \Ccal$, that are inside the $i$-th voxel. The likelihood function to update the semantic belief of the $j$-th class for the $i$-th voxel can be defined as follows.

\begin{equation}
p( c^{[j]} = +1 \mid Z) = \frac{1}{\lvert \Lcal^{[i]} \rvert} \sum_{s^{[k]}\in{\Lcal^{[i]}}} p(s^{[k]} \mid Z)
\end{equation}

\begin{figure}[t]
  \centering 
  \includegraphics[width=.99\columnwidth,trim={0cm 0cm 0cm 0cm},clip]{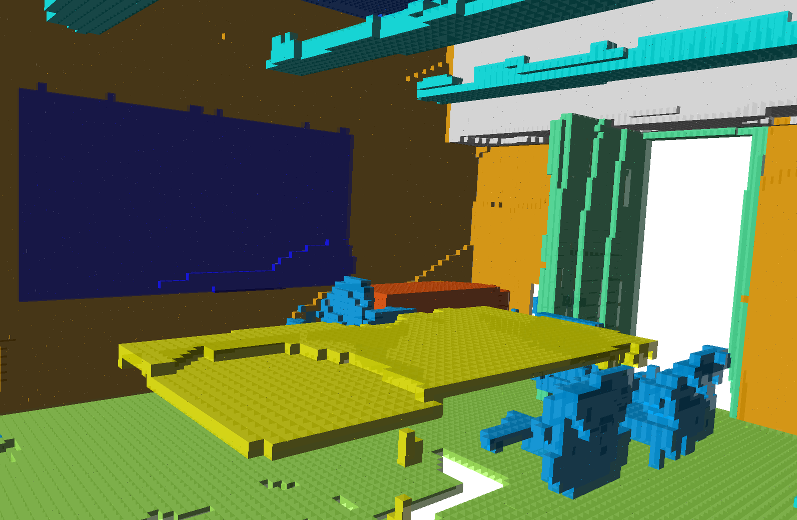}
  \caption{Semantic OctoMap built using the Stanford 2D-3D-Semantics Dataset~\cite{2017arXiv170201105A}, area 1, conference room 1. The semantic label observations are from the ground truth data. From the figure, the effect of discretization of the map and independent voxels inference are evident in the structural shape of the environment.}
  \label{fig:stanford}
\end{figure}

For the visualization purpose, we only visualize the occupied voxels with their corresponding hard labels which are computed by finding the label with the maximum semantic probability. Figure~\ref{fig:stanford} shows an example of the developed semantic OctoMap.

\section{Computational Complexity Analysis}
\label{sec:complexity}
FITC uses inducing points (active set) to base the computations on cross-covariances between training, test, and inducing points; hence, the computational cost is dominated by the matrix multiplication and reduced from $\bigO{n_t^3}$ to $\bigO{n_t n_u^2}$ where $n_u$ is the number of inducing points and $n_u \ll n_t$. The computational complexity of the GP semantic map is dominated by FITC approximation and is the same.

The computational complexity of semantic OctoMap is determined by octree data access complexity, which is \mbox{$\bigO{n_d} = \bigO{\log{n_n}}$}, where $n_n$ is the total number of nodes and $n_d$ is the depth of the octree data structure. As semantics fusion process queries the corresponding voxel on an octree for each measured point in a scan, the computational complexity of this part is $\bigO{n_p\log{n_n}}$, where $n_p$ is the number of points in the point cloud. It is worth mentioning that the maximum depth of octree is fixed in implementation; therefore, the practical computational complexity of the semantic fusion process is $\bigO{n_p}$.

\begin{figure*}[t!]
  \centering  
  \subfloat{
    \includegraphics[width=0.5\columnwidth]{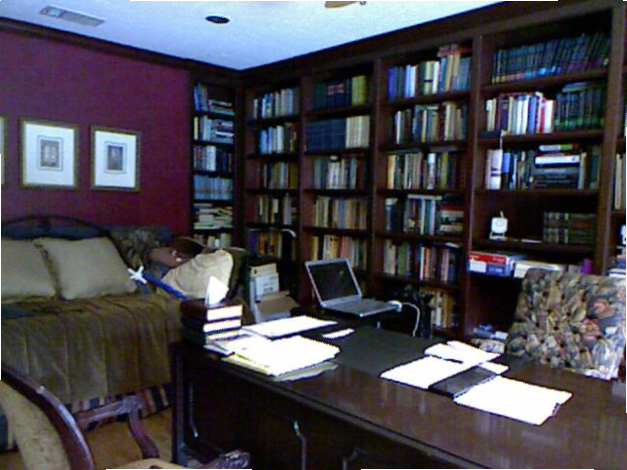}
    \label{fig:gt_a}
    }
  \subfloat{
    \includegraphics[width=0.5\columnwidth]{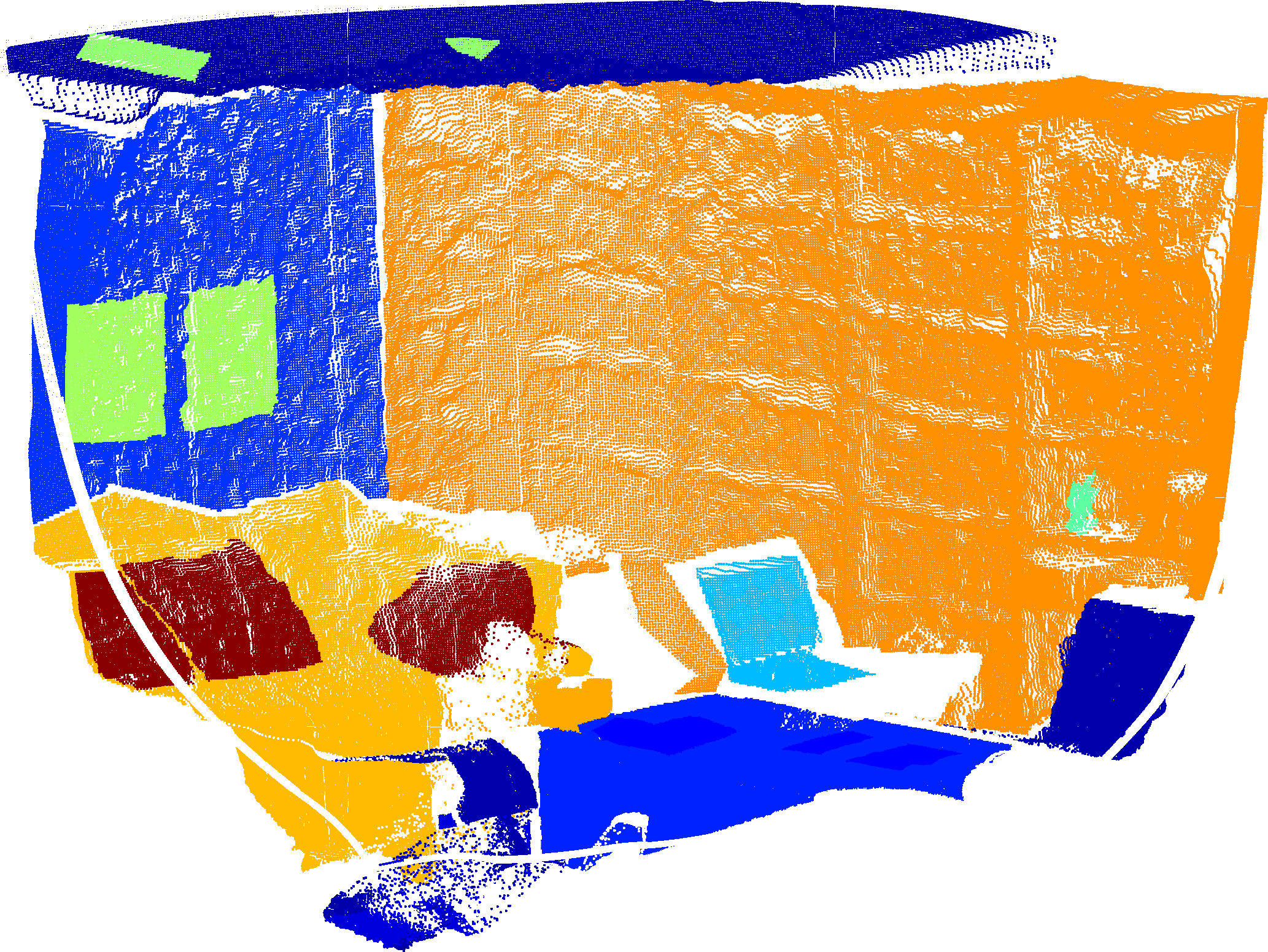}
    \label{fig:gtd_a}
    }
  \subfloat{
    \includegraphics[width=0.5\columnwidth]{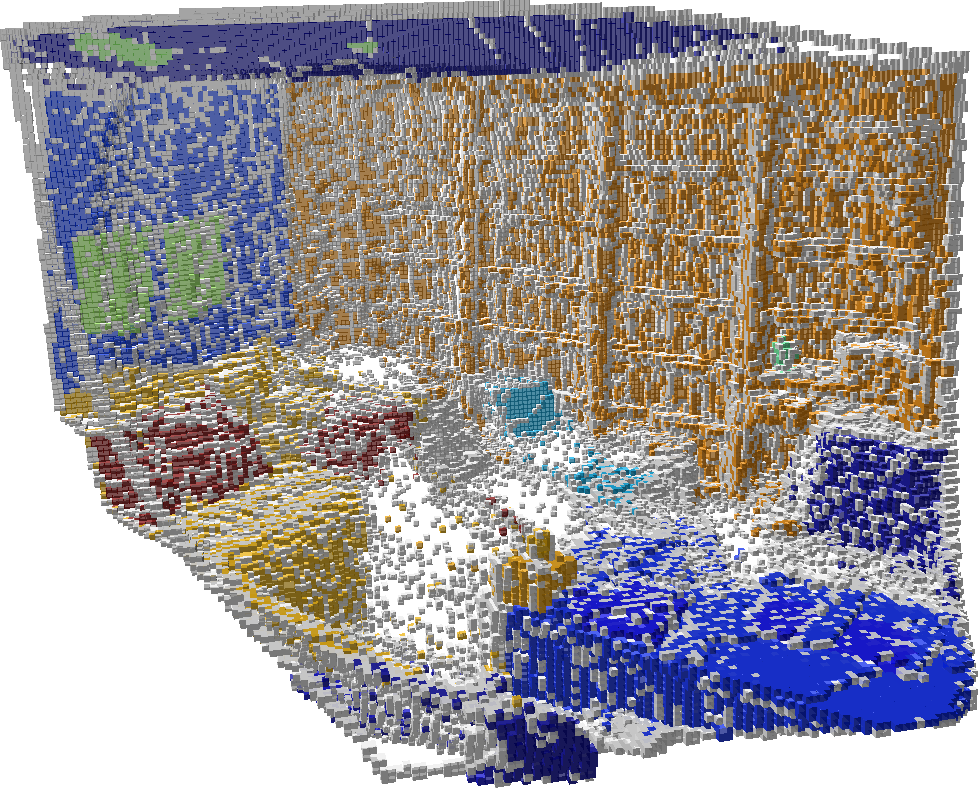}
    \label{fig:som_a}
    }
  \subfloat{
    \includegraphics[width=0.5\columnwidth]{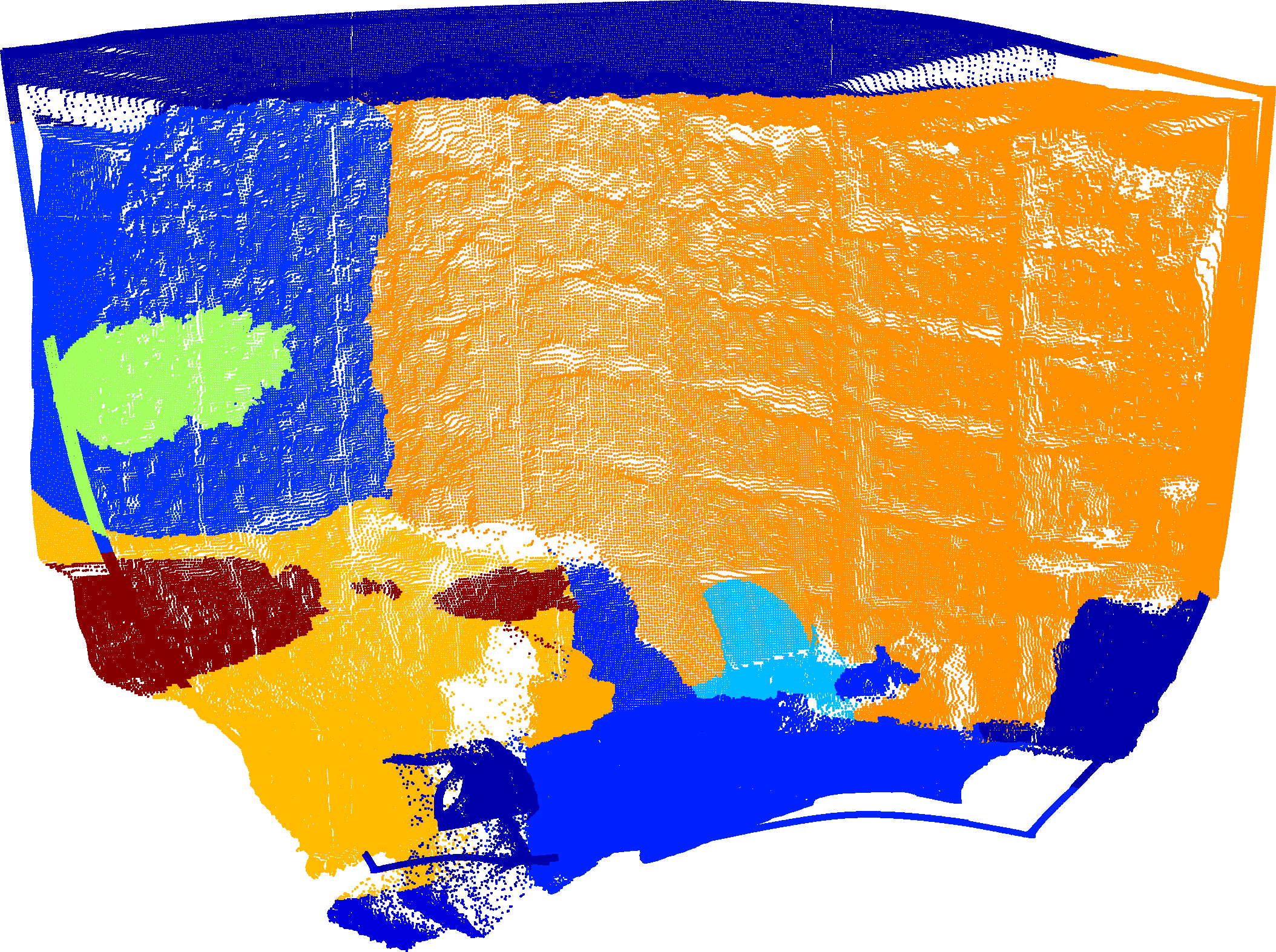}
    \label{fig:gpsm_a}
    }\\
  \subfloat{
    \includegraphics[width=0.5\columnwidth]{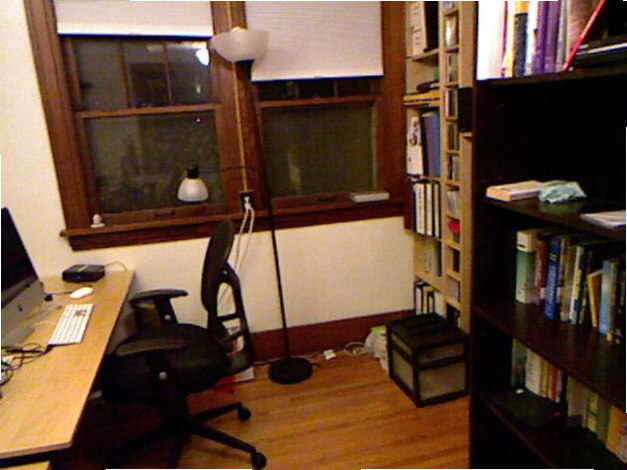}
    \label{fig:gt_b}
    }
  \subfloat{
    \includegraphics[width=0.5\columnwidth]{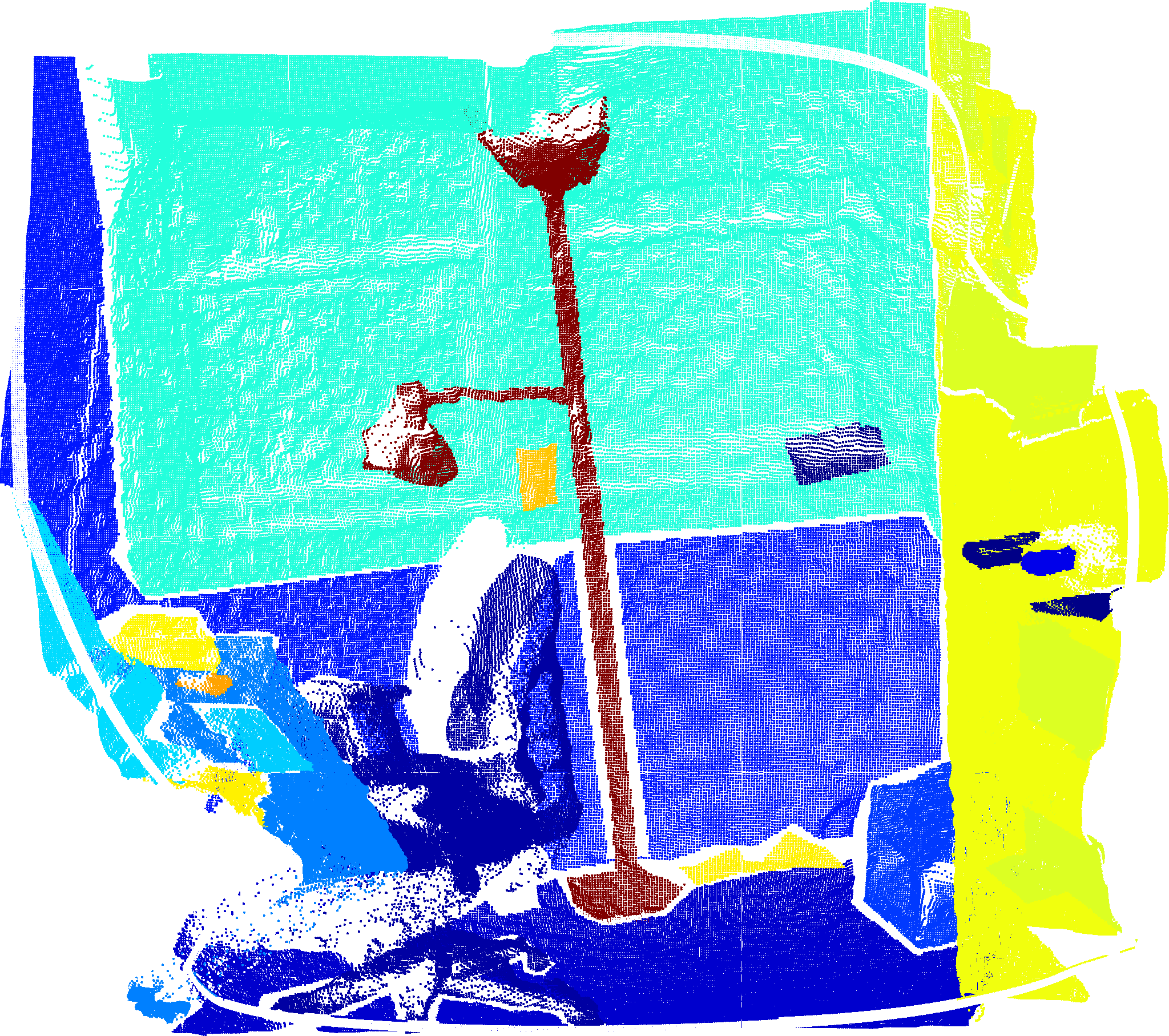}
    \label{fig:gtd_b}
    }
  \subfloat{
    \includegraphics[width=0.5\columnwidth]{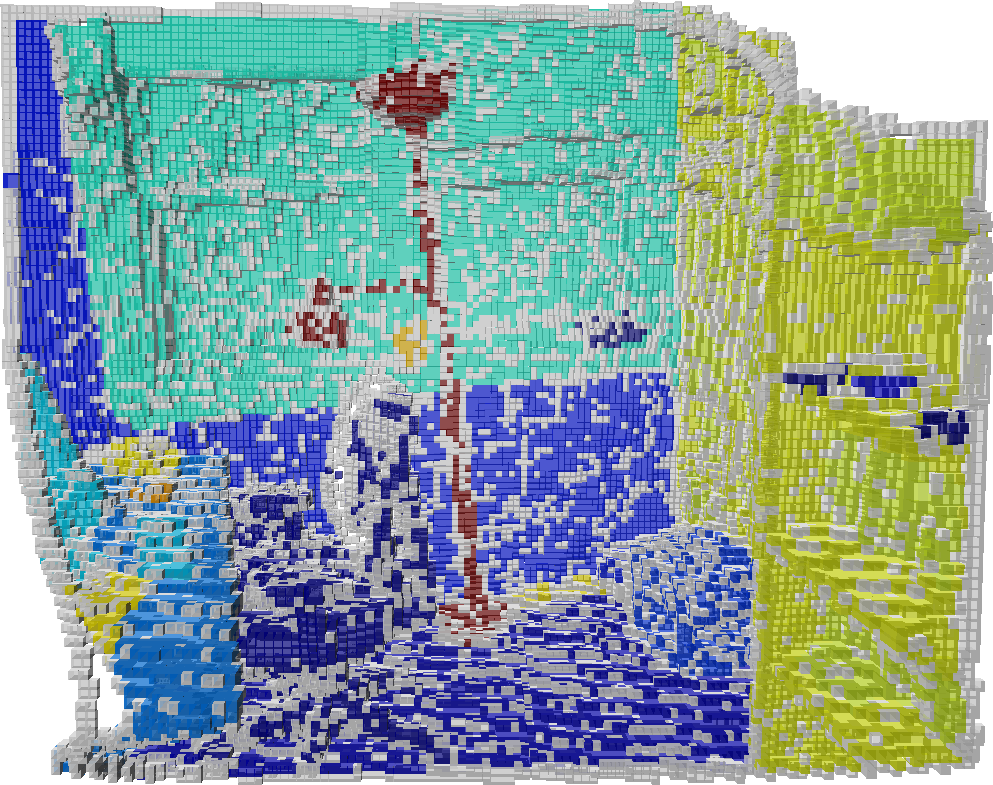}
    \label{fig:som_b}
    }
  \subfloat{
    \includegraphics[width=0.5\columnwidth]{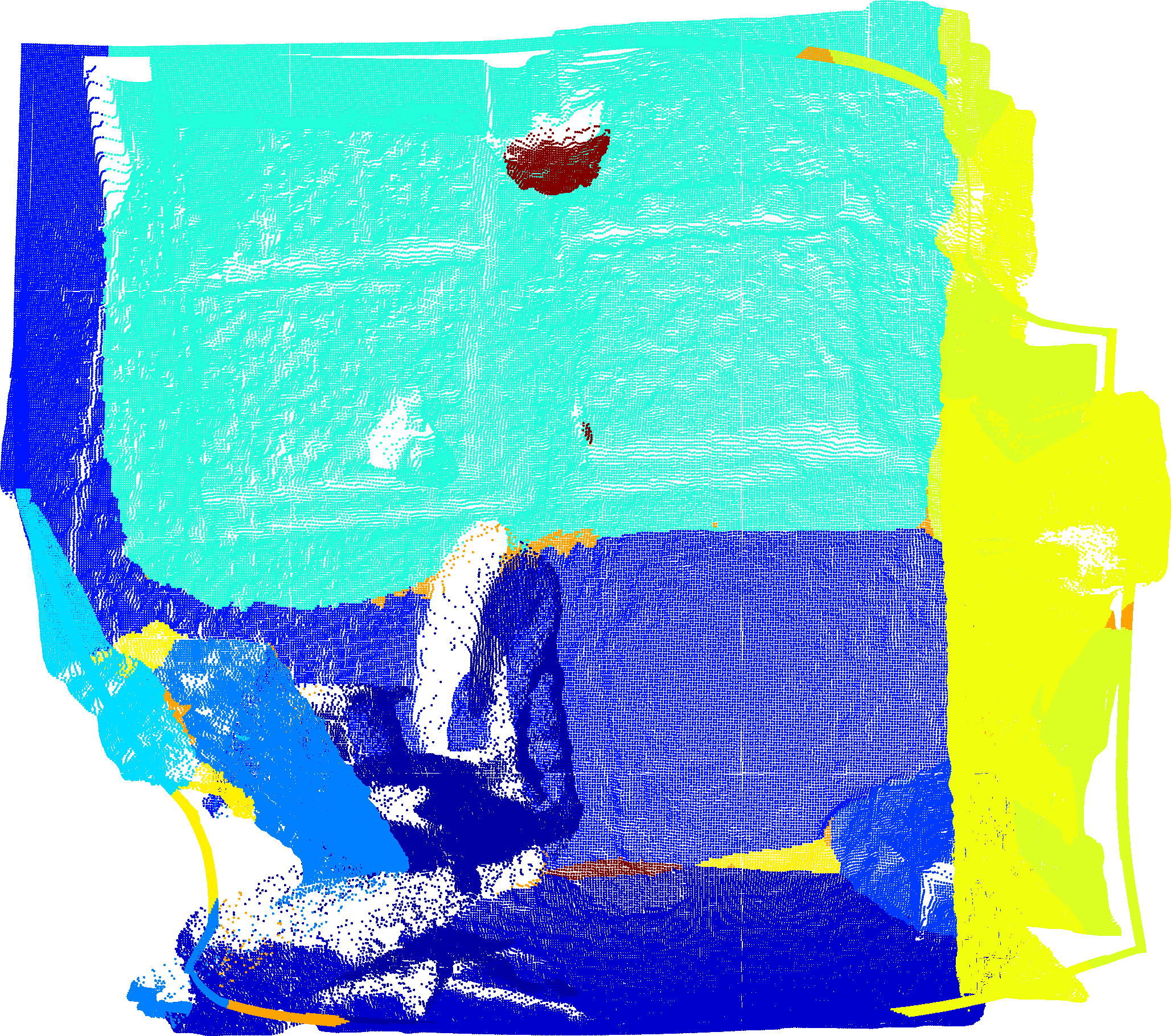}
    \label{fig:gpsm_b}
    }\\
  \subfloat{
    \includegraphics[width=0.5\columnwidth]{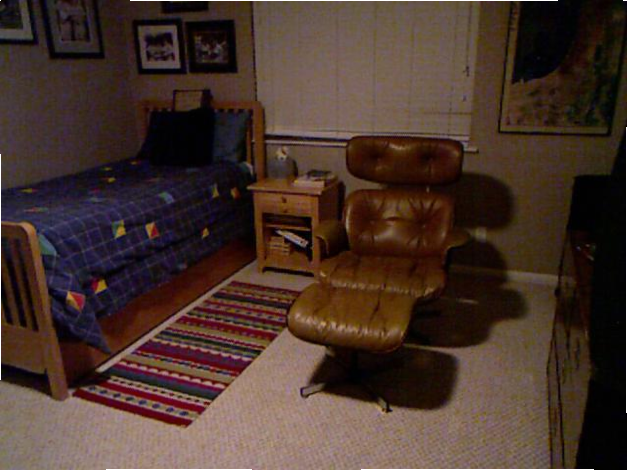}
    \label{fig:gt_c}
    }
  \subfloat{
    \includegraphics[width=0.5\columnwidth]{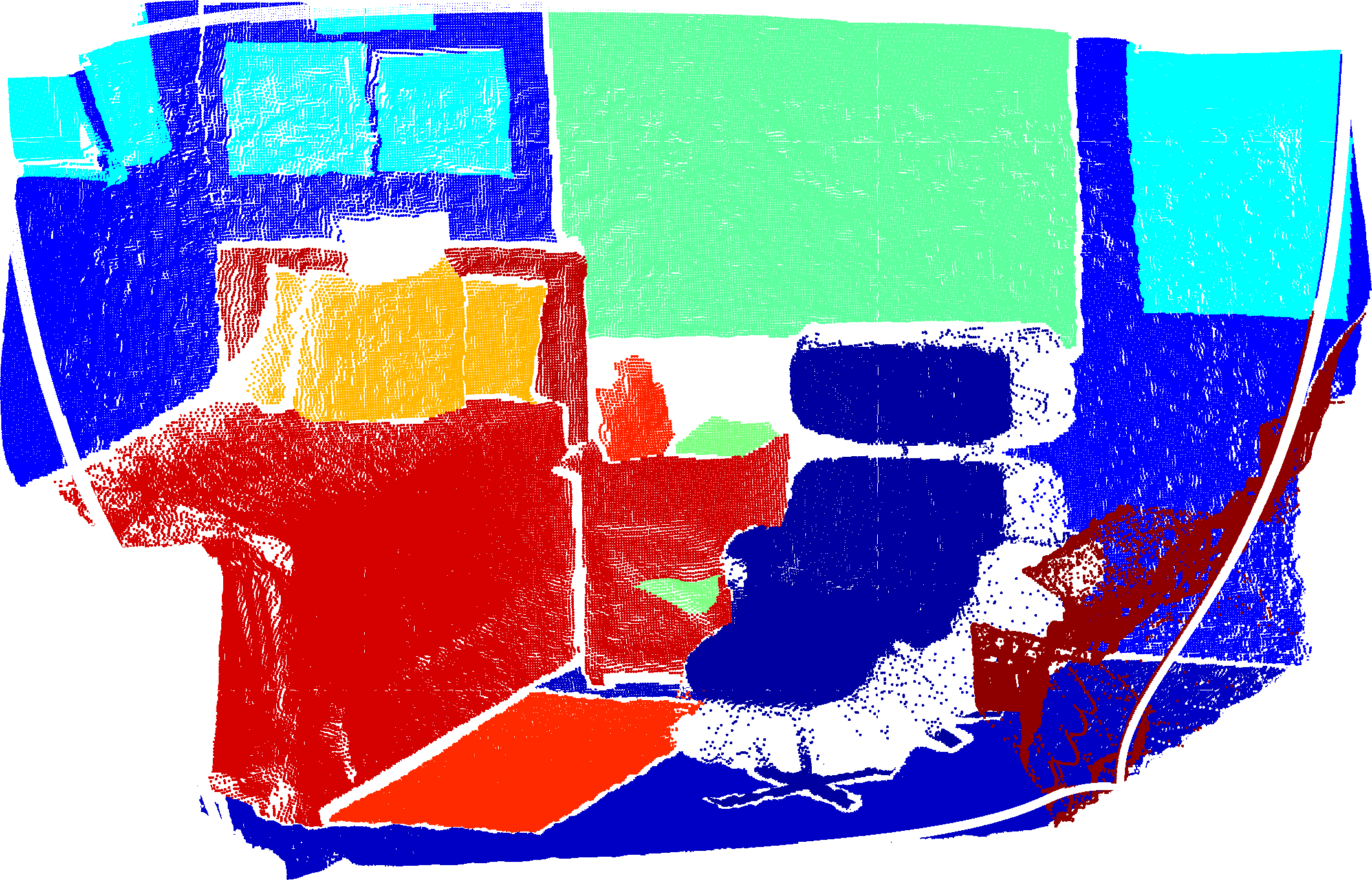}
    \label{fig:gtd_c}
    }
  \subfloat{
    \includegraphics[width=0.5\columnwidth]{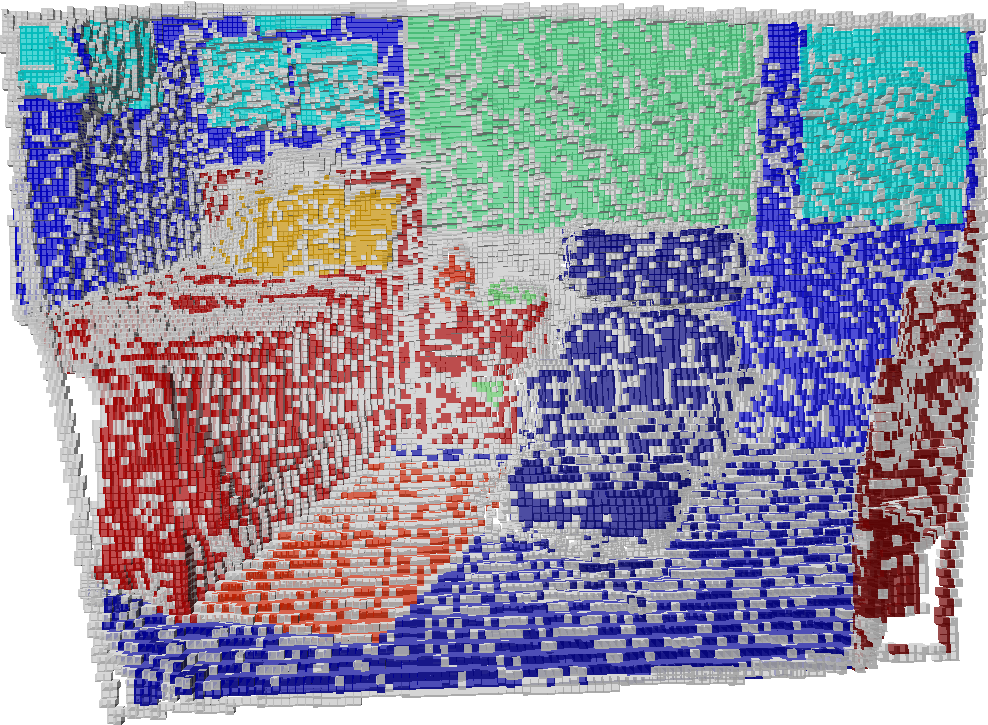}
    \label{fig:som_c}
    }
  \subfloat{
    \includegraphics[width=0.5\columnwidth]{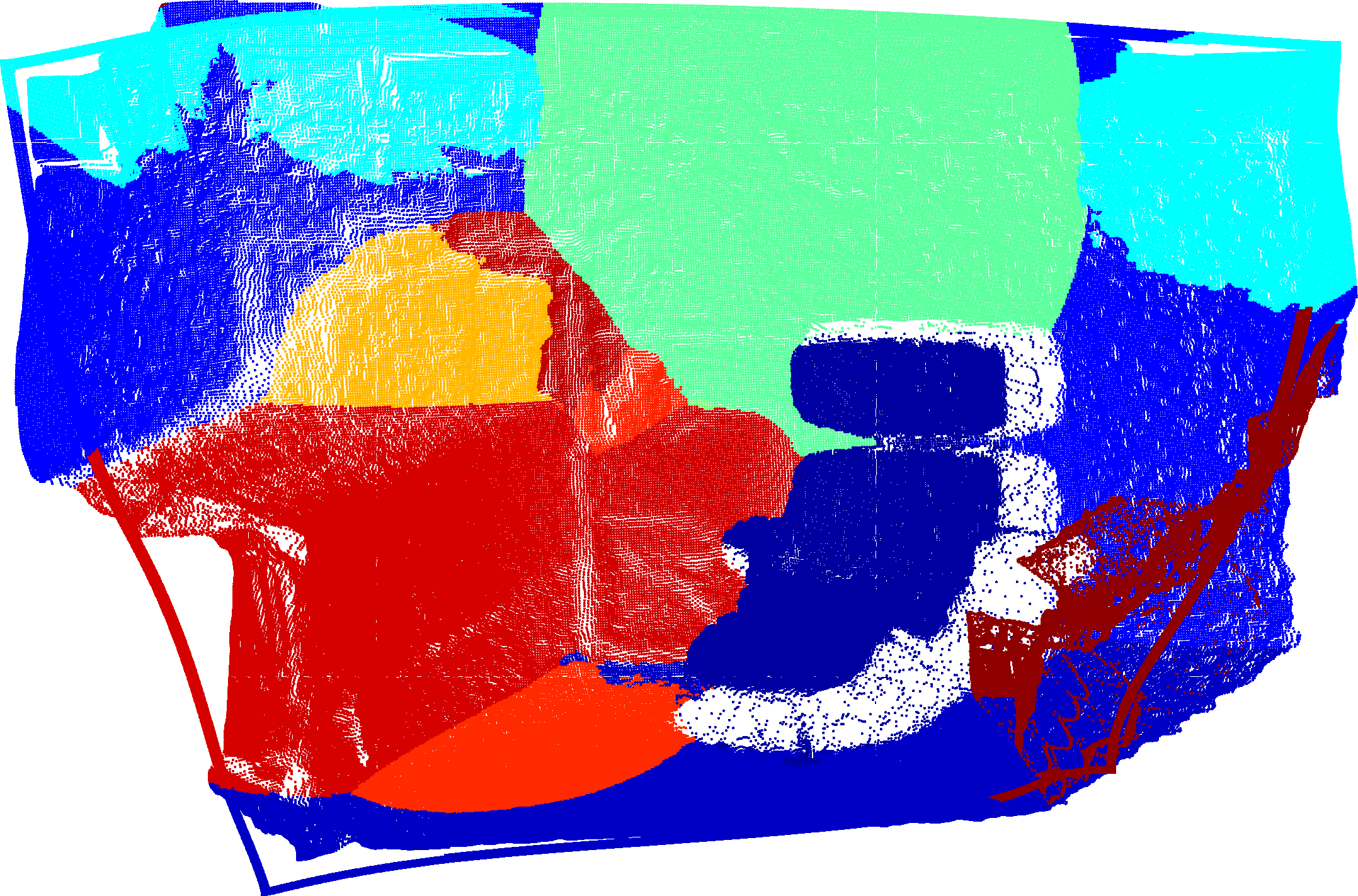}
    \label{fig:gpsm_c}
    }
  \caption{The figure shows GPSM and SOM results using sparse labeled point clouds. From left, each column respectively shows the RGB image, the point cloud with ground truth semantic labels before downsampling, the SOM, and the GPSM. The maps are built by uniformly down-sampling the original point cloud to one-third. The GP map can infer the missing labels and deal with sparse measurements through leaning the correlation between observations. For GPSM, the query points are the same as original points in the second column.}
  \label{fig:gt_maps}
\end{figure*}

\section{Results and Discussion}
\label{sec:results}

We now present mapping results using the proposed GP semantic map and the semantic OctoMap. We evaluate the proposed technique by comparing the mapping performance with that of the semantic OctoMap using NYU Depth V2 dataset~\cite{Silberman:ECCV12}. The GPSM is implemented in MATLAB using the GPML library~\cite{rasmussen2006gaussian} and the SOM is implemented in C++ using by developing the original OctoMap implementation~\cite{hornung2013octomap}. 

In the following, we explain the experimental setup and the performance criterion used for the comparison. We run two experiments; first, we study the effect of sparse measurements and missing labels by downsampling ground truth point clouds. Second, we use an image segmentation technique to label the entire point cloud to study the effect of misclassification (false positives) in observations.

\subsection{Experimental Setup and Evaluation Criterion}

The NYU Depth V2 dataset provides a large set of aligned RGB and depth images in indoor environments that are recorded by the Kinect sensor. The dataset also contains a subset of pixel-wise multi-class labeled images where there are unlabeled pixels in the annotated images for missing structures. We first convert RGBD scans to 3D point clouds with point-wise labels which serve as observation sets. Then we build the GPSM and the SOM. Note that, in this work, the map is referred to a local map built using a single scan (labeled point cloud). 
In the first experiment, point clouds are labeled using the ground truth semantic labels and downsampled by one-third to replicate sparse measurements and missing labels. In the second experiment, semantic labels are computed by the image segmentation technique SegNet~\cite{kendall2015bayesian}. The resolution of semantic OctoMap is set to $0.02 \m$ for all results. The observation set contains about $300,000$ points. While the GP training points are a much smaller subset of the original observation set (typically less than $5000$ points), the query points are all points in the original point cloud.

The evaluations include the comparison of mapping performance using the \emph{Area Under the receiver operating characteristic Curve} (AUC). The probability that the classifier ranks a randomly chosen positive instance higher than a randomly chosen negative instance can be understood using the AUC of the classifier; furthermore, the AUC is useful for domains with skewed class distribution and unequal classification error costs~\citep{fawcett2006introduction}. The AUC is originally a measure of the discriminability of a pair of classes. The extension of this method to the multi-class case is discussed in~\citet{hand2001simple}. In order to maintain the performance measure insensitive to class distribution and error costs, following our one-vs.-rest multi-class classification approach, we calculate an AUC for each classifier (GP) and then take the average AUC as the overall performance measure. Thus, the total AUC can be computed as follows.

\begin{equation}
\mathrm{AUC}_{\mathrm{total}} = \frac{1}{\lvert \Ccal \rvert} \sum_{j \in \Ccal} \mathrm{AUC}^{[j]}
\end{equation}

\begin{figure*}[t!]
  \centering  
  \subfloat{\includegraphics[width=.5\columnwidth]{nyu_v2_282_img}
  \label{fig:gt2}}
  \subfloat{\includegraphics[width=.5\columnwidth]{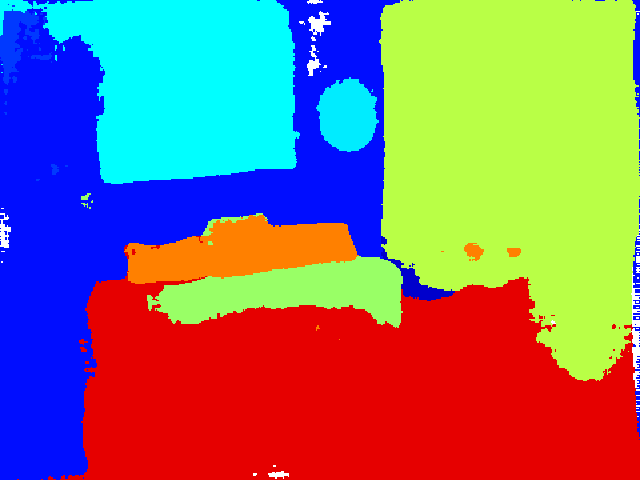}
  \label{fig:gtd2}}
  \subfloat{\includegraphics[width=.5\columnwidth]{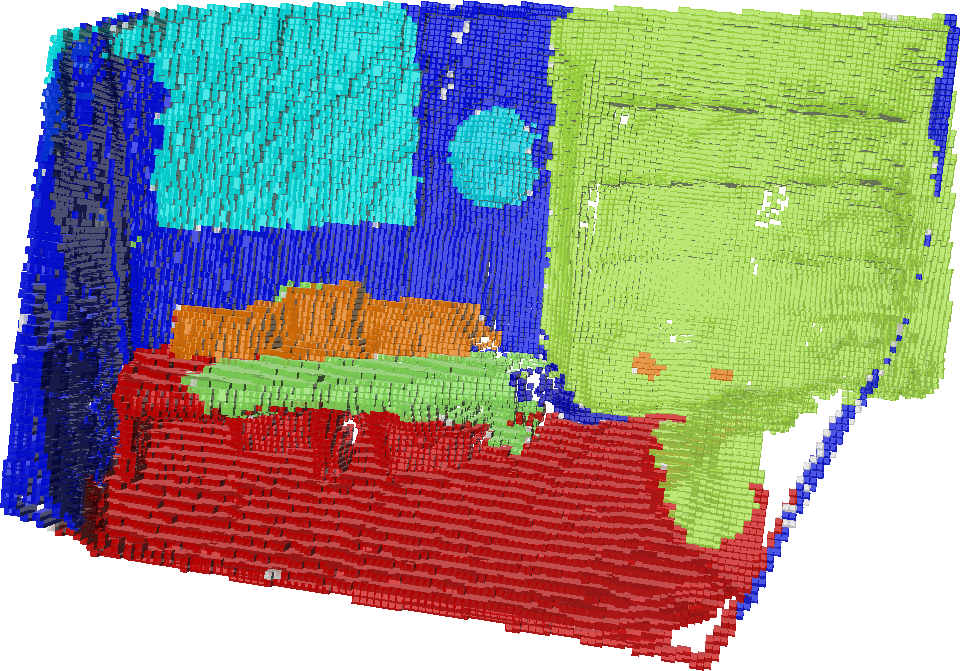}
  \label{fig:som2}}
  \subfloat{\includegraphics[width=.5\columnwidth]{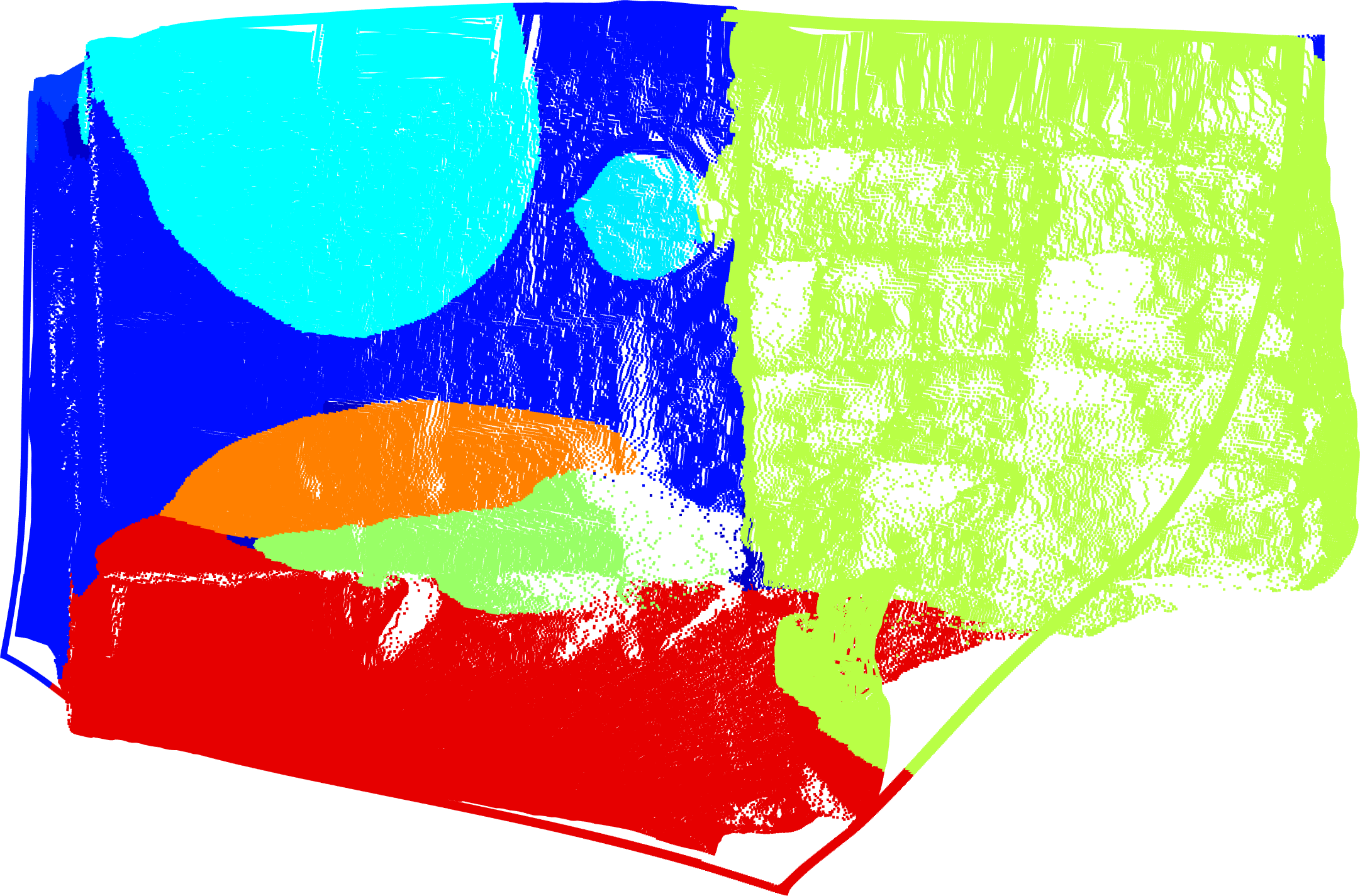}
  \label{fig:gpsm2}} \\
  \subfloat{\includegraphics[width=0.5\columnwidth]{nyu_v2_374_img}
    \label{fig:gt_a2}}
  \subfloat{\includegraphics[width=0.5\columnwidth]{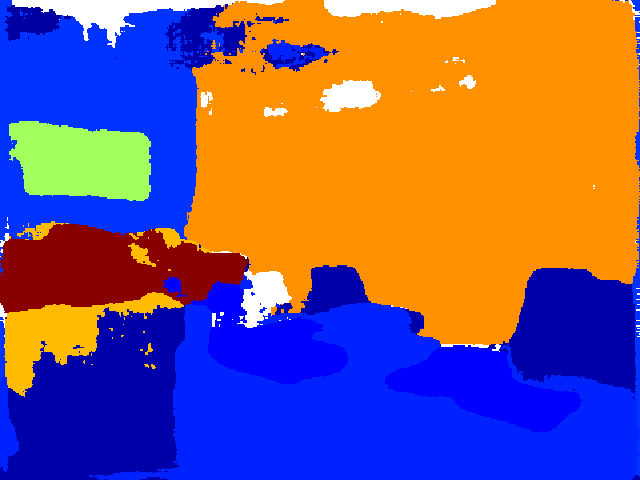}
    \label{fig:gtd_a2}}
  \subfloat{\includegraphics[width=0.5\columnwidth]{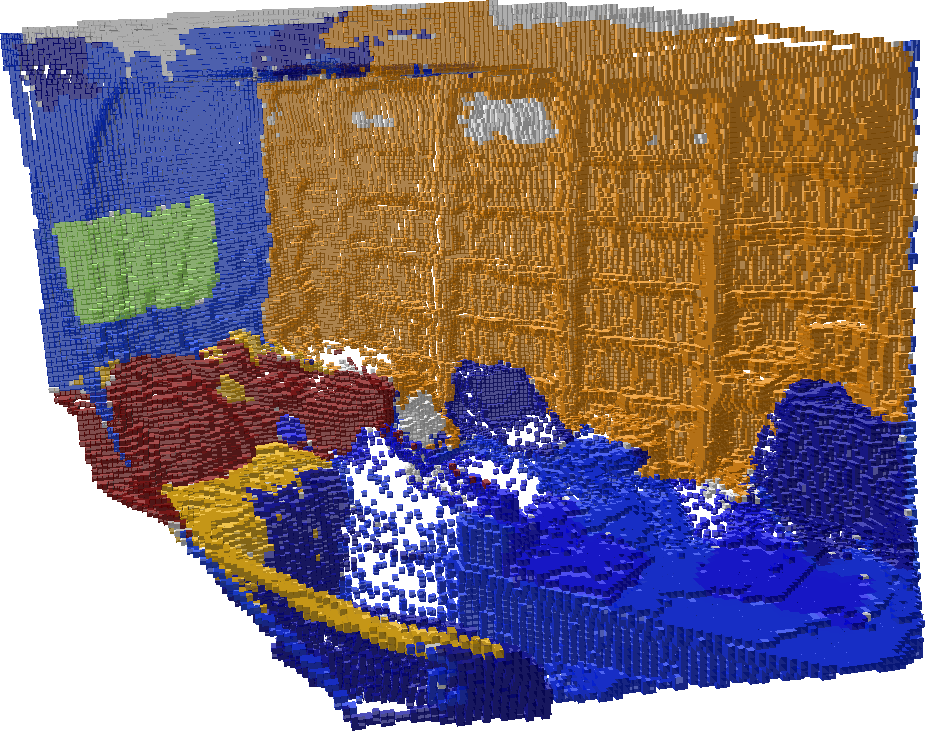}
    \label{fig:som_a2}}
  \subfloat{\includegraphics[width=0.5\columnwidth]{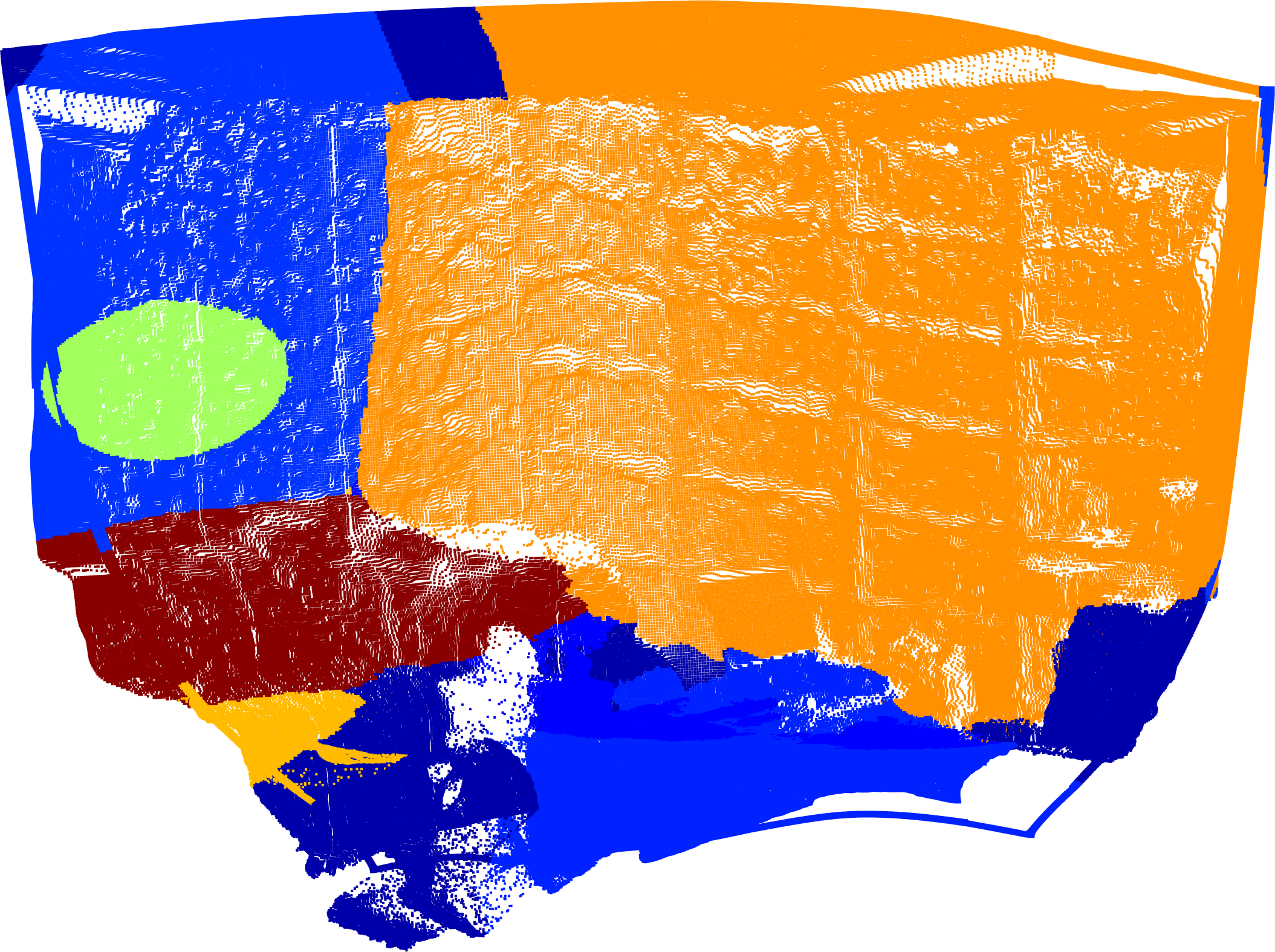}
    \label{fig:gpsm_a2}}\\
  \subfloat{\includegraphics[width=0.5\columnwidth]{nyu_v2_555_img}
    \label{fig:gt_b2}}
  \subfloat{\includegraphics[width=0.5\columnwidth]{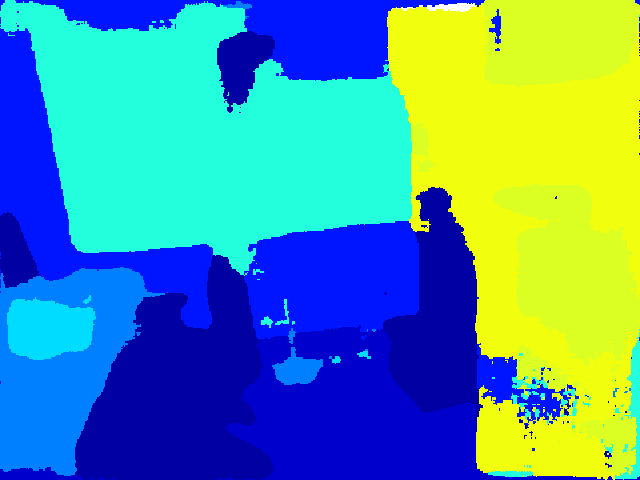}
    \label{fig:gtd_b2}}
  \subfloat{\includegraphics[width=0.5\columnwidth]{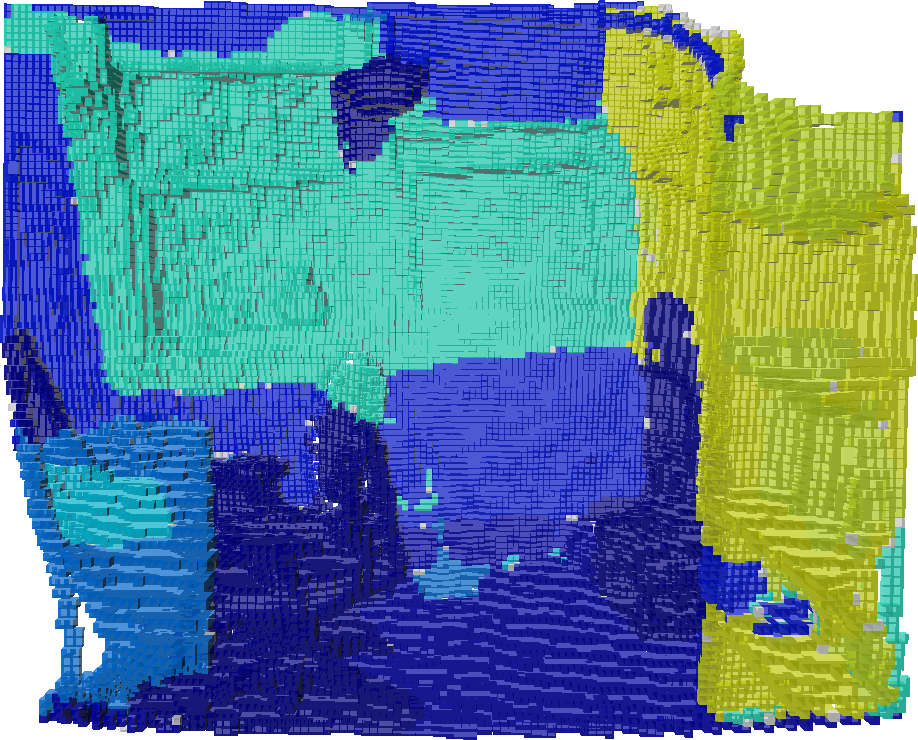}
    \label{fig:som_b2}}
  \subfloat{\includegraphics[width=0.5\columnwidth]{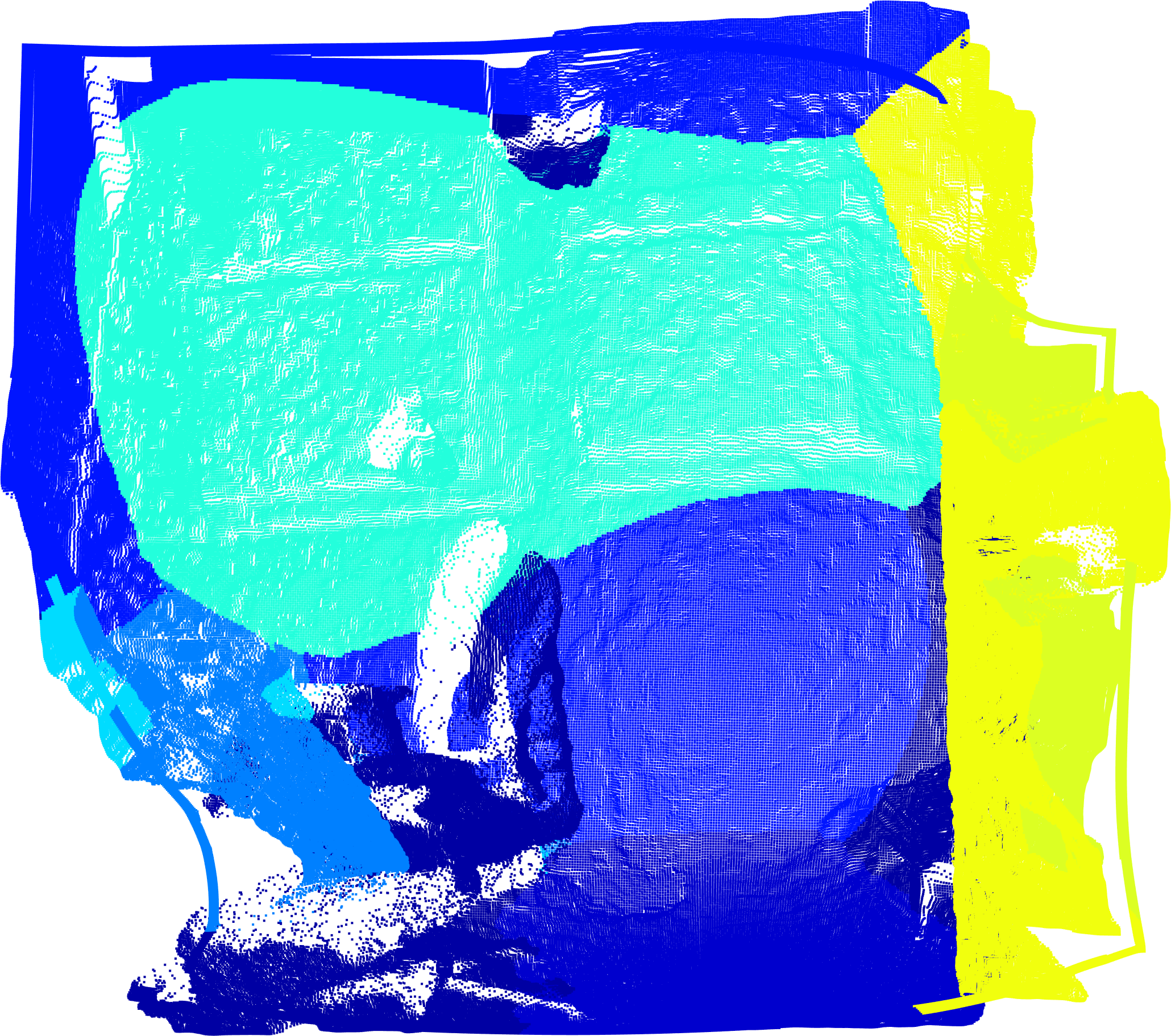}
    \label{fig:gpsm_b2}}\\
  \subfloat{\includegraphics[width=0.5\columnwidth]{nyu_v2_965_img}
    \label{fig:gt_c2}}
  \subfloat{\includegraphics[width=0.5\columnwidth]{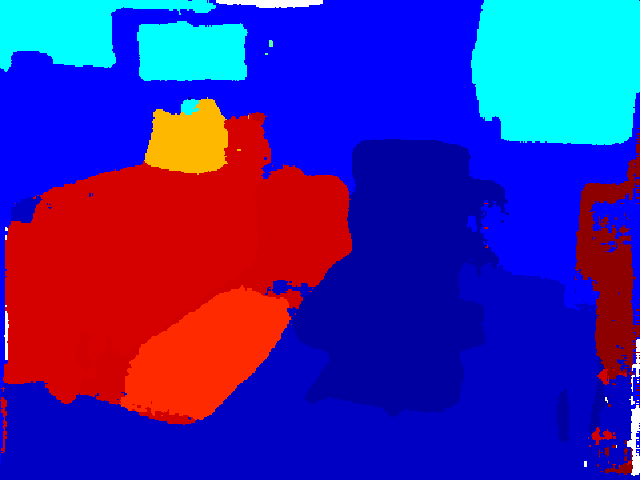}
    \label{fig:gtd_c2}}
  \subfloat{\includegraphics[width=0.5\columnwidth]{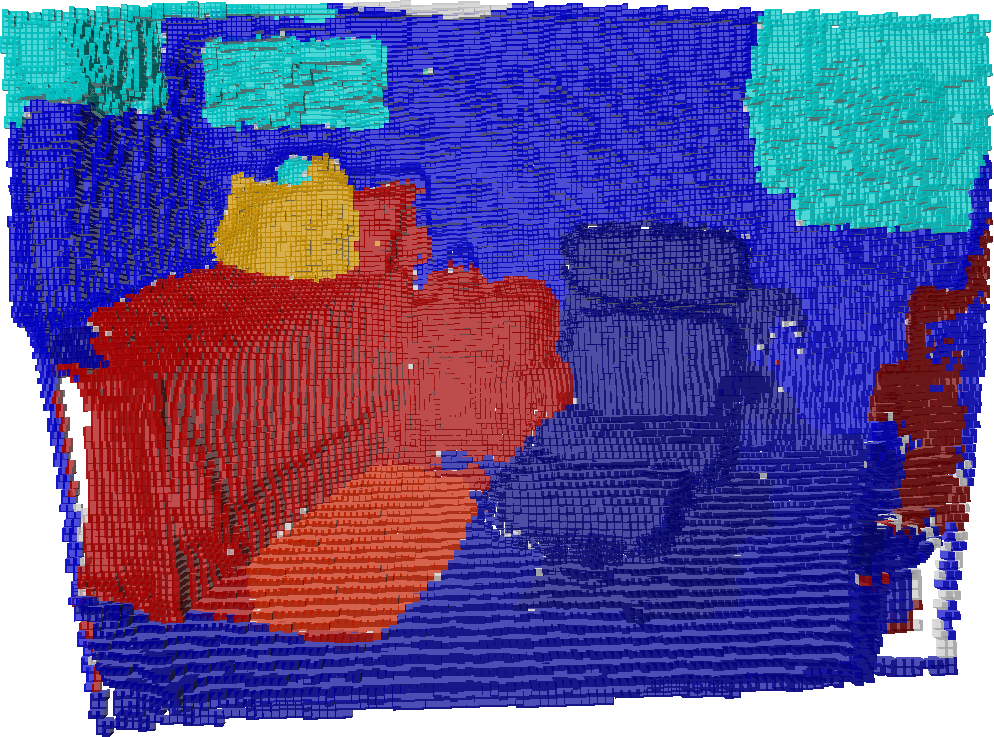}
    \label{fig:som_c2}}
  \subfloat{\includegraphics[width=0.5\columnwidth]{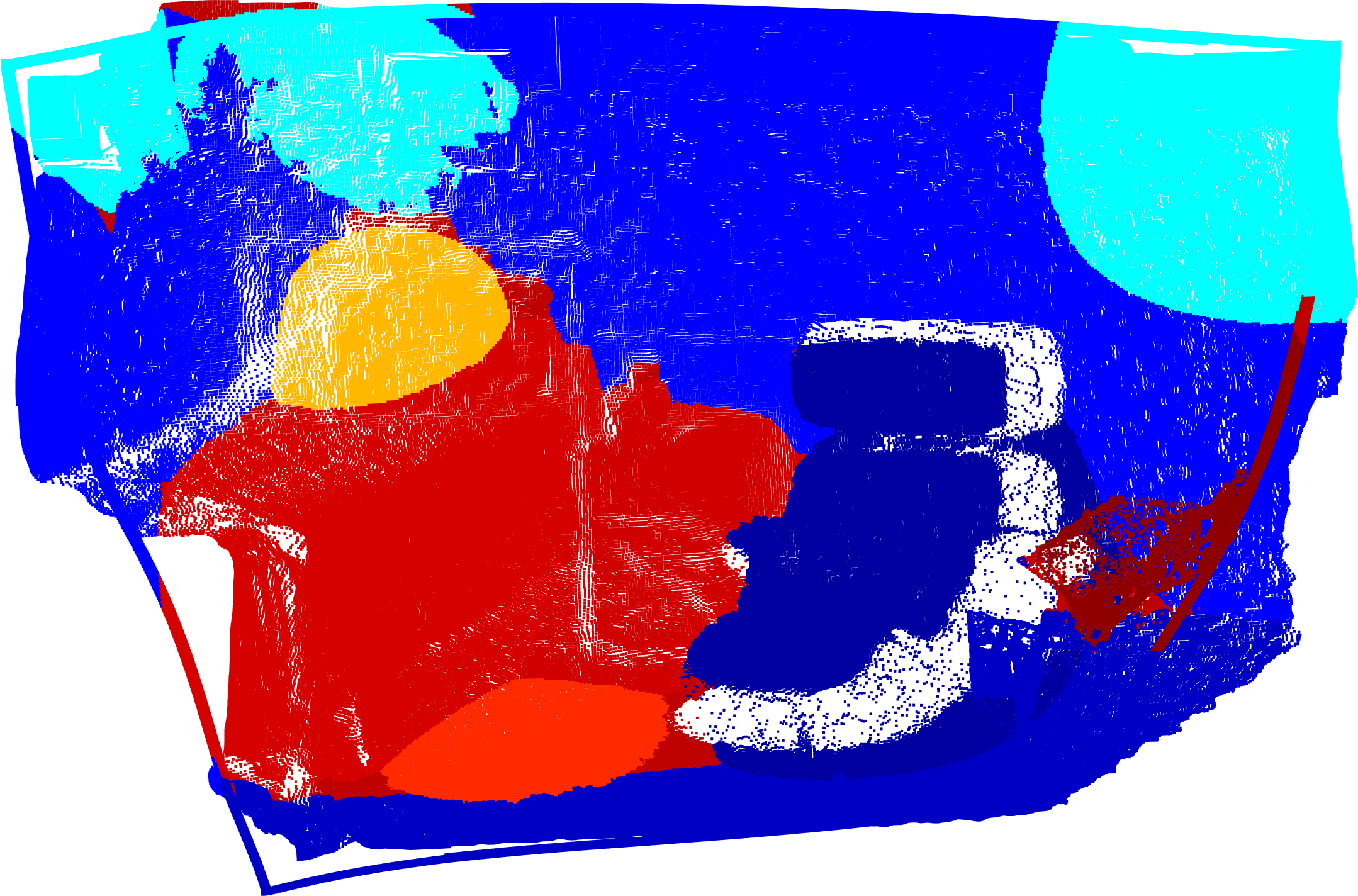}
    \label{fig:gpsm_c2}}
  \caption{The results of mapping under noisy and misclassified labels. From left, each column respectively shows the RGB image, the image segmentation using SegNet, the SOM, and the GPSM results. The false positives in the image segmentation are directly transferred into the SOM results, while the GPSM uses the spatial distance between points to infer the class labels based on correlations between map points.}
  \label{fig:segnet_maps}
\end{figure*}

\subsection{Sparse and Missing Measurements Effects}

In the first experiment, we use four samples from the dataset for map building. The point clouds are labeled using the ground truth data. We uniformly downsample the labeled point clouds by one-third to replicate sparse measurements. As a result, since the number of labeled points is reduced, the point cloud contains many unlabeled points which makes the mapping process challenging. Figures~\ref{fig:firstfig} and \ref{fig:gt_maps} show the results of GPSM and SOM using the selected frames. Table~\ref{tab:first_exp} shows the quantitative comparison between GPSM and SOM.

\begin{table}
\centering
\caption{The comparison of Gaussian processes semantic map and semantic OctoMap using the area under the receiver operating characteristic curve (AUC). The maps are built using downsampled point clouds from the NYU Depth V2 dataset~\citep{Silberman:ECCV12} and the corresponding frame numbers are shown in the table.}
\begin{tabular}{lcc}
\toprule
Frame Number	& \multicolumn{1}{c}{GP Semantic Map} 	& \multicolumn{1}{c}{Semantic OctoMap} \\
    			& \multicolumn{1}{c}{$\mathrm{AUC}_{\mathrm{total}}$} 				& \multicolumn{1}{c}{$\mathrm{AUC}_{\mathrm{total}}$} 				\\ \midrule
NYU V2 - 282 	& 0.9624		& 0.88525		\\ 
NYU V2 - 374 	& 0.9622		& 0.86251		\\ 
NYU V2 - 555 	& 0.9675		& 0.88648		\\ 
NYU V2 - 965 	& 0.9644		& 0.83569		\\ \bottomrule
\end{tabular}
\label{tab:first_exp}
\end{table}

From the result, the marginalization effect of the SOM on the map inference appears as many unlabeled voxels. Essentially, any voxel that does not contain any labeled points remains as unlabeled as there is no information available at that particular location. Intuitively, it is likely that neighboring locations share the same semantic class label and if any two points are spatially close, this chance can be even higher due to the present structural correlations in the environment. The lower mapping performance of the SOM, while dealing with sparse measurements, approves this claim. 

The GP semantic map places a joint distribution on the observations and query points. This high-dimensional approach enables the method to infer the semantic class labels continuously. Furthermore, the available information in observations are captured in the GP covariance function which effectively models the correlation between map points. The GPSM is not limited to a fixed resolution and provides a probabilistic prediction and any desired location. The GPSM performs on all samples equally well with minor misclassified regions in each map. The maps are visualized using the class label with the maximum probability.

\subsection{Mapping under Noisy and Misclassified Labels}
In the second experiment, we use SegNet~\cite{kendall2015bayesian} to generate semantic labels. We first perform the RGB image segmentation and then assign each pixel's label to its corresponding point in the point cloud. The mapping results are computed using the same four examples from the previous experiment, and the performance of each method is evaluated using the ground truth labeled point clouds in Figure~\ref{fig:gt_maps}. Figure~\ref{fig:segnet_maps} shows the results of the image segmentation (second column from left), the SOM (third column from left), and the GPSM (fourth column from left). Table~\ref{tab:second_exp} shows the quantitative comparison between GPSM and SOM. The challenge in this test is the presence of false positive labels in the observations and, as expected, both methods achieve lower mapping performance on the same set of data. Note that in this experiment the point clouds are not downsampled; however, the GP training set is a subset of observations as explained in the previous subsection.

The SOM directly uses the available points within a voxel to infer the voxel labels. Therefore, any misclassified label is transferred to the map. However, the GPSM infers the class labels based on the available correlations in the observation sets. Specifically, in the presented examples, the spatial correlations reduce the number of the misclassified points in the final map. For example, two adjacent pixels in the image space can have a similar label while their spatial distance can be large. Such cases are trivially handled in the GPSM framework since a large spatial distance implies insignificant correlation.

\subsection{Discussion and Limitations}
Since we seldom have sufficient information about a process, minimization of the NLML is a useful approach as it provides flexibility for model selection. However, the problem of minimizing NLML is an ill-posed problem and using maximum likelihood estimate we can often find a local minimum. Therefore, it is always possible that the solution suffers from overfitting, especially if the number of hyperparameters is large~\citep{rasmussen2006gaussian}.

The common way for map representation in robotics is using a dense set of points with a particular distribution that is possibly suitable for navigation tasks. However, there is no restriction for any other desired representation such as approximate belief representations in~\citet{charrow2014approximate}, that can be useful for other applications such as predictions. In fact, it is shown in~\citet{maani2017ijrr} that by modeling the underlying process as GPs, one can perform information-theoretic planning using nonparametric information gain with possibility to handle the state estimate uncertainty~\citep{maaniwgpom}.

In heteroscedastic processes, noise is state dependent. Heteroscedastic Gaussian process regression~\citep{goldberg1998regression,kersting2007most} can be an alternative to model the structural correlation more accurately. However, to model the prediction uncertainty, they require a second GP in addition to the GP governing the noise-free output value. The computational cost is roughly twice that of the standard GP~\citep{ko2009gp,titsias2011variational}. To avoid increasing the computational time in the proposed mapping technique, we do not consider the heteroscedasticity in training data; however, applying these techniques to the problem at hand is an interesting direction to follow.

\section{Conclusion and Future Work} 
\label{sec:conclusion}

In this paper, we developed the Gaussian processes semantic map for 3D semantic map representations. We formulated the problem as a multi-class classification by defining the map spatial coordinates and semantic class labels as the input and output of the process, respectively. The proposed GP semantic map learns the structural and semantic correlation from measurements rather than resorting to assumptions, and exploits the spatial correlation (and possibly any additional non-spatial correlation) between map points for inference. In particular, the proposed map inference can infer missing labels and deal with sparse measurements, is continuous and queries can be made at any desired locations, and can deal with false positives better.

While, in this work, we only considered the spatial coordinates as the input, GPSM is agnostic to the input dimensions and can handle an arbitrary number of non-spatial dimensions. Future work includes extension of the proposed method to an incremental form and addition of other available dimensions such as color and intensity.

\begin{table}[t]
\centering
\caption{The comparison of Gaussian processes semantic map and semantic OctoMap using the area under the receiver operating characteristic curve (AUC). The maps are built using noisy image segmentation labels produced by SegNet. The figures for both methods are lower than the first experiment due to the presence of false positive labels in the observations.}
\begin{tabular}{lcc}
\toprule
Frame Number	& \multicolumn{1}{c}{GP Semantic Map} 	& \multicolumn{1}{c}{Semantic OctoMap} \\
    			& \multicolumn{1}{c}{$\mathrm{AUC}_{\mathrm{total}}$} 				& \multicolumn{1}{c}{$\mathrm{AUC}_{\mathrm{total}}$} 				\\ \midrule
NYU V2 - 282 	& 0.7192		& 0.68402		\\ 
NYU V2 - 374 	& 0.7625		& 0.73552		\\ 
NYU V2 - 555 	& 0.7032		& 0.66881		\\ 
NYU V2 - 965 	& 0.8612		& 0.77456		\\ \bottomrule
\end{tabular}
\label{tab:second_exp}
\end{table}


\small
\bibliographystyle{plainnat}
\bibliography{references}

\end{document}